\documentclass[10pt,twocolumn,letterpaper]{article}

\usepackage[pagenumbers]{iccv} 

\definecolor{iccvblue}{rgb}{0.21,0.49,0.74}
\usepackage[pagebackref,breaklinks,colorlinks,allcolors=iccvblue]{hyperref}
\usepackage{paralist}

\usepackage{amsthm}
\usepackage{caption}
\usepackage{subcaption}
\usepackage{multirow}
\usepackage{bbm}
\usepackage{mathtools}
\usepackage{mathabx}
\usepackage{duckuments}
\usepackage{amssymb}
\usepackage{enumitem}
\usepackage{algorithm}
\usepackage{algpseudocode}
\usepackage{wrapfig}
\usepackage{float}
\usepackage{makecell}
\usepackage{adjustbox}
\usepackage{booktabs, multirow} 
\usepackage{soul}
\usepackage{colortbl} 
\usepackage{changepage,threeparttable} 

\usepackage{amsmath,amsfonts,bm}
\usepackage[most]{tcolorbox}

\def\1{\bm{1}}

\def\rvepsilon{{\mathbf{\epsilon}}}

\def\rvd{{\mathbf{d}}}

\def\rvn{{\mathbf{n}}}

\def\rvv{{\mathbf{v}}}

\def\rvx{{\mathbf{x}}}

\def\rmD{{\mathbf{D}}}

\def\rmN{{\mathbf{N}}}

\def\rmX{{\mathbf{X}}}

\DeclareMathAlphabet{\mathsfit}{\encodingdefault}{\sfdefault}{m}{sl}
\SetMathAlphabet{\mathsfit}{bold}{\encodingdefault}{\sfdefault}{bx}{n}

\def\gD{{\mathcal{D}}}
\def\gE{{\mathcal{E}}}

\usepackage[most]{tcolorbox}

\newtcolorbox{mblock}[1]
{
  colframe     = gray,
  coltitle     = black,
  colbacktitle = lightgray!50!white,
  colback      = white,
  title        = #1,
  fonttitle    = \bfseries,
  arc          = 0mm,
  left         = 2pt,
  right        = 2pt,
}

\newtcolorbox{rblock}[1]
{
  colframe     = green,
  coltitle     = gray,
  colbacktitle = lightgray!50!white,
  colback      = white,
  title        = \ifthenelse{\isempty{#1}}
                  {Remark}
                  {Remark: #1},
  fonttitle    = \bfseries,
  arc          = 0mm,
  left         = 2pt,
  right        = 2pt,
}

\PassOptionsToPackage{table}{xcolor}
\usepackage{xcolor}                 
\definecolor{NVIDIAGreen}{HTML}{76B900}

\usepackage{pifont}
\newcommand{\cmark}{\ding{51}}  
\newcommand{\xmark}{\ding{55}}  
\definecolor{gentlepink}{RGB}{255, 105, 130}     
\definecolor{gentleviolet}{RGB}{145, 100, 255}   
\definecolor{gentleblue}{RGB}{100, 160, 210}     

\hypersetup{
    urlcolor=NVIDIAGreen
}

\def\paperID{} 
\def\confName{}
\def\confYear{}

\title{\vspace{-8mm}GeoMan:  Temporally Consistent Human Geometry Estimation using \\ Image-to-Video Diffusion}
\author{
Gwanghyun Kim$^{1,2*}$, Xueting Li$^1$, Ye Yuan$^1$, Koki Nagano$^1$, Tianye Li$^1$, 
\\ Jan Kautz$^1$, Se Young Chun$^2$,\stepcounter{footnote} Umar Iqbal$^{1}$ \\
   \normalfont $^1$NVIDIA \quad  
  \normalfont $^2$Seoul National University  \\
\url{https://research.nvidia.com/labs/dair/geoman}
}

\begin{document}
\maketitle
\begin{abstract}
Estimating accurate and temporally consistent 3D human geometry from videos is a challenging problem in computer vision. 
Existing methods, primarily optimized for single images, often suffer from temporal inconsistencies and fail to capture fine-grained dynamic details. 
To address these limitations, we present GeoMan, a novel architecture designed to produce accurate and temporally consistent depth and normal estimations from monocular human videos. 
GeoMan addresses two key challenges: the scarcity of high-quality 4D training data and the need for metric depth estimation to accurately model human size. 
To overcome the first challenge, GeoMan employs an image-based model to estimate depth and normals for the first frame of a video, which then conditions a video diffusion model, reframing video geometry estimation task as an image-to-video generation problem. 
This design offloads the heavy lifting of geometric estimation to the image model and simplifies the video model’s role to focus on intricate details while using priors learned from large-scale video datasets. 
Consequently, GeoMan improves temporal consistency and generalizability while requiring minimal 4D training data. 
To address the challenge of accurate human size estimation, we introduce a root-relative depth representation that retains critical human-scale details and is easier to be estimated from monocular inputs, overcoming the limitations of traditional affine-invariant and metric depth representations. 
GeoMan achieves state-of-the-art performance in both qualitative and quantitative evaluations, demonstrating its effectiveness in overcoming longstanding challenges in 3D human geometry estimation from videos.

\end{abstract}

{
  \renewcommand{\thefootnote}{\fnsymbol{footnote}}
  \footnotetext[1]{Work done during an internship at NVIDIA.}
}

\vspace{-1mm}
\section{Introduction}
\vspace{-2mm}
\label{sec:introduction}
\begin{figure}[!t]
    \centering
    \includegraphics[width=0.95\columnwidth]{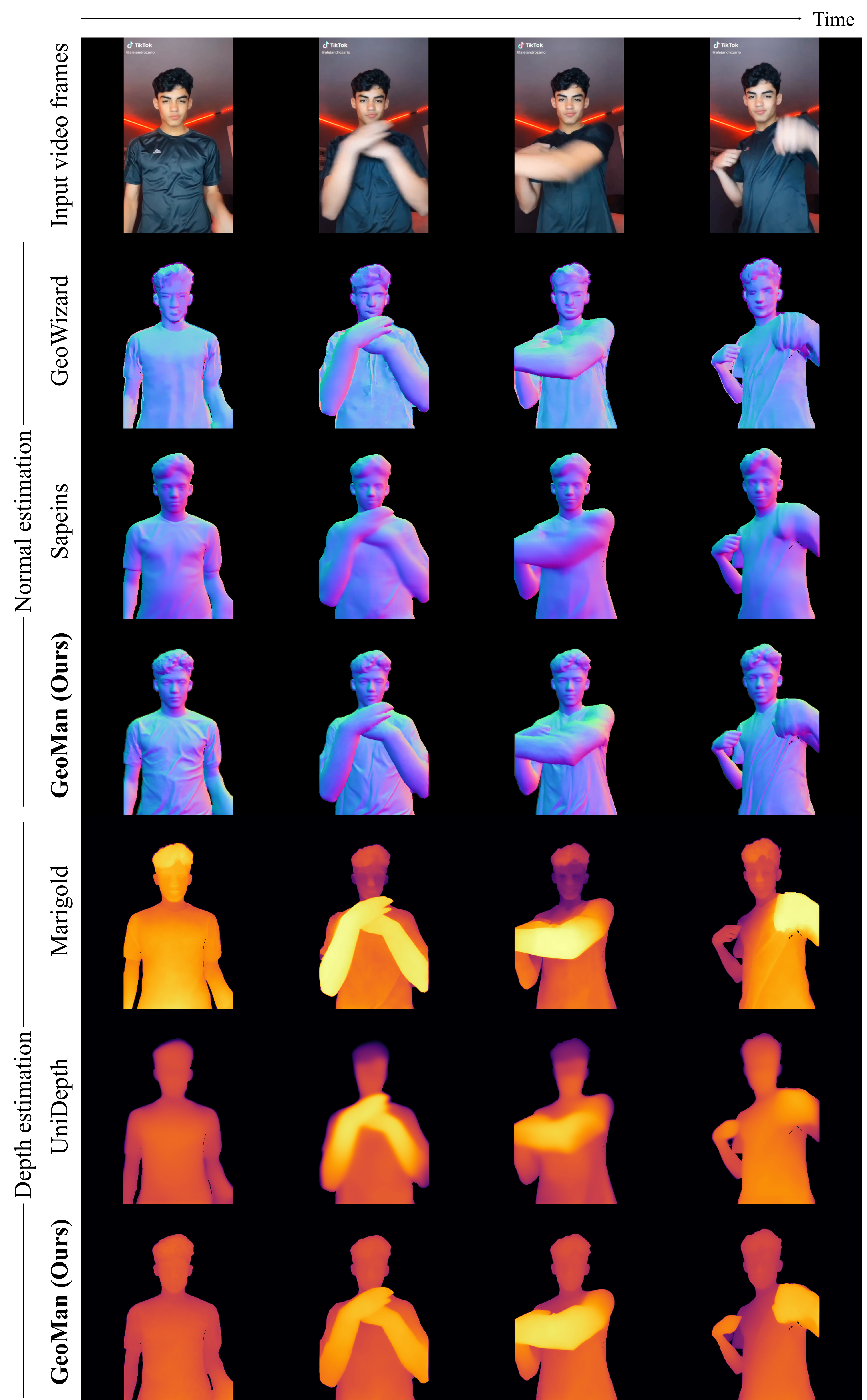}
    \vspace{-0.5em}
    \caption{GeoMan provides accurate and temporally stable geometric predictions for human videos, surpassing existing methods.}
    \label{fig1}
    \vspace{-1.5em}
\end{figure}

Estimating human 3D geometry~\cite{saito2019pifu, saito2020pifuhd, xiu2022econ} from images or videos has enabled numerous applications in digital human and virtual reality. 
While significant progress~\cite{Jafarian_2021_CVPR_TikTok, depthanything, depth_anything_v2} has been made, most methods~\cite{khirodkar2024sapiens,marigold,he2024lotus} focus on single-image geometric estimation. 
When applied to videos, these methods often struggle with temporal inconsistency, causing visible flickering artifacts and missing fine dynamic details such as subtle facial deformations, clothing wrinkles, and hair movements. These details are crucial for applications involving human motion, as shown in Fig.~\ref{fig1}
This work aims to estimate highly accurate and temporally consistent depth and normal from monocular human videos.

This is a highly challenging task that presents two difficulties.
First, the scarcity of high-quality 4D data poses a significant challenge to training accurate geometry estimation models. 
Recent works~\cite{marigold, fu2024geowizard} address this limitation by repurposing text-to-image diffusion models for image-based geometry estimation. These methods modify the diffusion denoiser architecture to intake both the input image and geometric information, followed by fine-tuning on image-geometry pairs. 
Extending these approaches to video geometry estimation introduces additional challenges due to the higher dimensionality and complexity of videos. While some methods~\cite{hu2024-DepthCrafter,yang2024depthanyvideo} have explored adapting video diffusion models to video depth estimation, they often require a substantial amount of high-quality video-depth sequences. For instance, DepthCrafter~\cite{hu2024-DepthCrafter} curates a large-scale dataset including approximately 200K real and synthetic video-depth pairs. Consequently, this strategy cannot be naively extended to human video geometry estimation, as high-quality 4D human data is difficult to collect.
The second challenge lies in selecting an appropriate depth representation. Depth estimation from monocular inputs is inherently ambiguous. Most methods resort to predicting affine-invariant depth, which normalizes the depth map using the minimum and maximum depth value for each image or video.
While affine-invariant depth simplifies the learning objective and encourages the models to focus on local geometric details, it sacrifices essential size and scale information, particularly for human subjects, resulting in distorted 3D geometry and temporal flickering as presented in Fig.~\ref{fig:depth_sota_comparison}. 
Recent methods~\cite{bhat2023zoedepth,bochkovskii2024depthpro,piccinelli2024unidepth,hu2024metric3d} have attempted to predict metric depth to address these issues; however, they often do so at the expense of spatial details.

To address these challenges, we introduce \textit{GeoMan}, a novel approach for temporal consistent and accurate depth and normal estimation from monocular human videos.
To reduce dependency on 4D training data, we decompose the complex video geometry estimation task into image geometry estimation and image-to-video synthesis. This allows us to maximally exploit the prior in image-to-video diffusion models by introducing minimal changes to the diffusion model architecture. As a result, we maintain its powerful representation capabilities and generalization ability while adapting it for our specific task with limited 4D training data.
To achieve this, we first train an image-based model to estimate depth and normal maps of static images. 
This model is then used to predict geometric information of the first frame of a video, which serves as conditioning input for an image-to-video diffusion model. 
The video model subsequently infers temporally consistent geometric information for the entire video. 
Furthermore, GeoMan unifies depth and normal estimation by leveraging a single video diffusion model. This is achieved by simply switching the conditioning image, enabling simultaneous training for both modalities (depth and normal) and enhancing generalization across diverse scenarios.
Second, to preserve human scale information in depth estimation, GeoMan introduces a root-relative depth representation tailored for human-centric geometry estimation. 
This representation retains essential metric-scale details while discarding only global translation, facilitating accurate local geometric details estimation.
When necessary, absolute translation can be efficiently recovered using state-of-the-art human pose estimation models (e.g., ~\cite{yuan2022glamr, wang2024tram}), enabling absolute metric depth estimation.
Unlike conventional per-frame or per-video normalization techniques that hinder metric depth recovery and degrade temporal stability by computing independent scale factors per frame or per video, our approach enables consistent and stable depth estimation over time. Our contributions are:
\begin{compactitem}
    \item We propose GeoMan, which repurposes image-to-video diffusion models for temporally consistent depth and normal estimation in human videos, even when trained on limited data. 
    \item  We introduce a root-relative depth representation that preserves critical human size information, enabling metric depth estimation and 3D reconstruction.
    \item Extensive evaluations show that GeoMan surpasses baselines in accuracy and temporal stability, while effectively generalizing to in-the-wild human videos.
\end{compactitem}

\vspace{-1mm}
\section{Related Works}
\vspace{-2mm}
\noindent \textbf{Image-Based Geometry Estimation.}
Image-based depth and normal estimation has advanced significantly in recent years.
By leveraging mixed datasets and powerful diffusion models~\cite{blattmann2023stable}, state-of-the-art approaches such as MiDaS~\cite{ranftl2020towards}, Depth-Anything~\cite{depthanything, depth_anything_v2}, Marigold~\cite{marigold}, and GeoWizard~\cite{fu2024geowizard} achieve high accuracy and impressive generalization.
In parallel, a line of research has focused on human-centric geometry estimation. This approach utilizes either implicit surface regression~\cite{saito2019pifu,saito2020pifuhd, alldieck2022phorhum,han2023high} or parametric body models (e.g., SMPL~\cite{SMPL-X:2019})~\cite{loper2015smpl,zheng2020pamir, huang2020arch, xiu2022econ, xiu2022icon, zhang2023sifu}. Recently, HDNet~\cite{Jafarian_2021_CVPR_TikTok} proposes a generalizable self-supervision method that leverages large-scale unlabeled human videos for training. 
Additionally, Sapiens~\cite{khirodkar2024sapiens} demonstrates strong performance by pretraining on large-scale human image datasets. 
Despite these advancements, these image-based methods often struggle with temporal consistency and fine geometric details when applied to video geometry estimation, as shown in Fig.~\ref{fig:normal}.

\begin{figure*}[!t]
    \centering
    \includegraphics[width=0.92\textwidth]{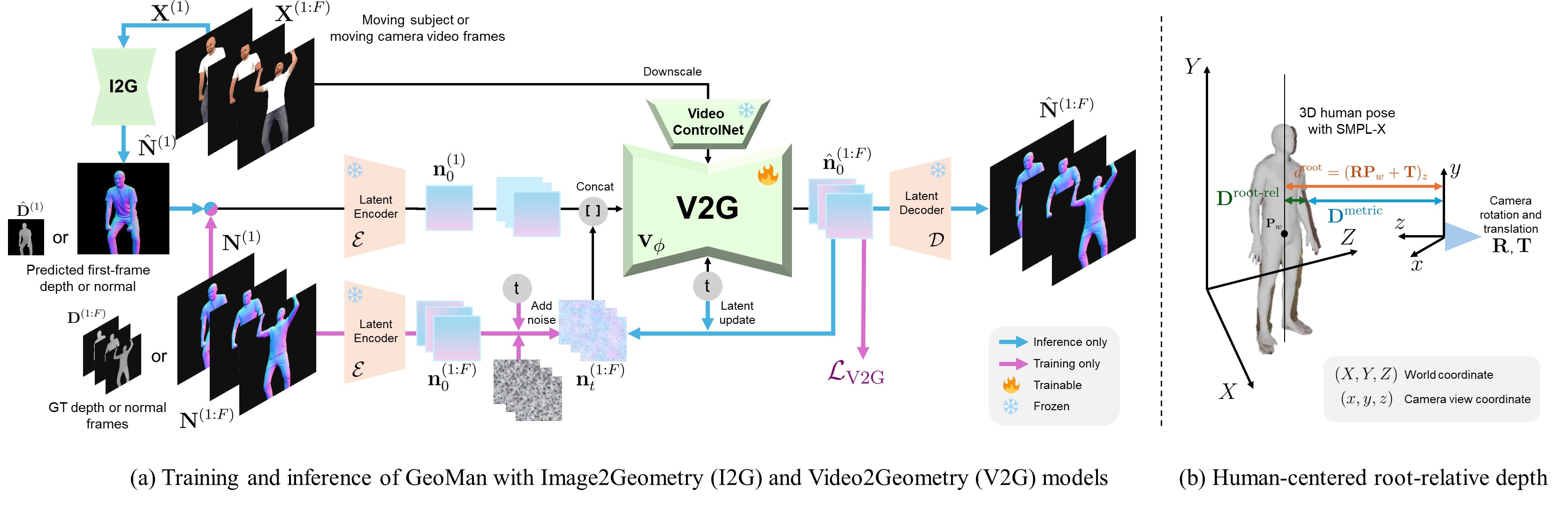}
    \vspace{-1.em}
    \caption{Overview of GeoMan: (a) Given a video sequence $\mathbf{X}^{(1:F)}$ as input, we first use I2G to estimate the normal or depth of the first frame $\mathbf{X}^{(1)}$. This initial prediction is then used to condition the V2G model, which generates predictions for the entire input sequence. GeoMan seamlessly handles both depth and normal estimation tasks using the same model weights, requiring only a replacement of the input condition for the first frame. (b) We propose a human-centered root-relative depth representation, which retains the human scale information and enables better temporal modeling.}
    \label{fig:overview}
    \vspace{-1.5em}
\end{figure*}

\noindent \textbf{Video Geometry Estimation.}
To enhance temporal consistency in video geometry estimation, 
one approach involves test-time optimization~\cite{Luo-VideoDepth-2020,kopf2021rcvd,zhang2021consistent,chen2019self}, which refines predictions using optical flow or camera poses. While effective, this post-processing step is often computationally expensive. 
Meanwhile, methods~\cite{hu2024-DepthCrafter,yang2024depthanyvideo} developed for generic object video geometry estimation often yield unsatisfying precision when applied to human videos.
In this paper, we aim at predicting both temporally stable and accurate geometry from human videos.

\noindent \textbf{Depth Representations.}
Existing depth estimation methods primarily utilize affine-invariant or metric depth representation.
Affine-invariant depth methods~\cite{marigold, fu2024geowizard, he2024lotus, hu2024-DepthCrafter, yang2024depthanyvideo} normalize depth maps using the minimum and maximum depth values of each image, effectively capturing relative depth structures while remaining robust to scale variations. 
However, the lack of absolute scale information limits their applicability to tasks requiring metric measurements.
Metric depth estimation methods~\cite{bhat2023zoedepth, piccinelli2024unidepth, yin2023metric3d, bochkovskii2024depthpro} recover depth in real-world units, enabling precise spatial reasoning. 
Yet, these methods often sacrifice fine geometric details due to the large numeric range of metric depth values, leading to less accurate local structure preservation. 
To address these limitations, we propose a root-relative depth representation that integrates human characteristics into depth representation. As shown in Fig.~\ref{fig:depth-comparison}, our approach enables metric depth estimation while preserving local geometric details, which cannot be achieved simultaneously by affine-invariant or metric depth estimation methods. 

\noindent \textbf{Image and Video diffusion model.}
Diffusion models~\cite{ho2020denoising, sohl2015deep} have emerged as powerful tools for generating high-quality images~\cite{ramesh2022hierarchical, ruiz2023dreambooth, zhang2023adding, podell2023sdxl, saharia2022photorealistic, ding2022cogview2} and videos~\cite{wang2023modelscope, he2022latent, chai2023stablevideo, chen2024videocrafter2, blattmann2023stable, bar2024lumiere, sora, chen2023videocrafter1, wang2024magicvideo, zhou2022magicvideo, 2023i2vgenxl} from text descriptions. 
Advancements such as Stable Video Diffusion (SVD)~\cite{blattmann2023stable}, and Video Diffusion Models (VDM)~\cite{ho2022video} generate videos with temporal consistency, while SORA~\cite{sora} recently set new benchmarks in photorealism and long-duration coherence.
To enhance controllability, few approaches integrate additional conditioning signals such as depth, skeletons, or optical flow~\cite{esser2023gen1, wang2024videocomposer, hu2023animateanyone, lin2024ctrl}. 
Notably, Ctrl-Adapter~\cite{lin2024ctrl} introduces an adapter module that enables fine-grained control over image and video diffusion models by leveraging pretrained ControlNet~\cite{zhang2023controlnet}, allowing users to specify motion trajectories, structural constraints, and content attributes.
We propose a novel framework that repurposes both image and controllable video diffusion models to generate accurate, temporal and multi-view consistent human geometry. 

\vspace{-1mm}
\section{Preliminaries}
\vspace{-2mm}
\subsection{Geometry Estimation via Image Diffusion}
\vspace{-1mm}
\label{sec:preliminary_image}
Recent methods have improved monocular depth~\cite{marigold,fu2024geowizard} and normal estimation~\cite{fu2024geowizard} by leveraging image diffusion models~\cite{rombach2022high} pre-trained on large-scale datasets.
The pioneering work Marigold~\cite{marigold} frames monocular depth estimation as a conditional image generation task, where the goal is to learn the conditional distribution $p(\rmD~|~\rmX)$ over depth $\rmD \in \mathbb{R}^{H \times W}$, using the RGB image $\rmX \in \mathbb{R}^{3 \times H \times W}$ as the condition.
To improve computational efficiency, Marigold adopts the Latent Diffusion Model (LDM)~\cite{rombach2022high} architecture which leverages a latent representation learned via a Variational Autoencoder (VAE)~\cite{kingma2013auto}. The encoder $\gE$ maps the depth map $\rmD$ to the latent representation: $\rvd = \gE(\rmD)$ while the decoder $\gD$ reconstructs the depth map from the latent: $\hat{\rmD} = \gD(\rvd)$. The RGB input $\rmX$ is similarly encoded as $\rvx = \gE(\rmX)$. 
Next, in the forward diffusion process, Gaussian noise is added to the depth latent feature $\rvd_0 := \rvd$ over $T$ steps: $\rvd_t = \sqrt{\bar{\alpha}_t} \rvd_0 + \sqrt{1 - \bar{\alpha}_t} \rvepsilon$,
where $\rvepsilon \sim \mathcal{N}(0, I)$, $\bar{\alpha}_t := \prod_{s=1}^{t}{(1 - \beta_s)}$, and $\{\beta_1, \ldots, \beta_T\}$ defines the variance schedule. 
The model is optimized by minimizing the denoising diffusion objective:
\begin{align}
\mathcal{L} = \mathbb{E}_{\rvd_0, \rvepsilon \sim \mathcal{N}(0, I), t \sim \mathcal{U}(T)} \left\| \rvepsilon - \rvepsilon_\theta(\rvd_t, \rvx, t) \right\|^2_2.
\vspace{-3mm}
\end{align}
To ensure stable performance, the $v$-objective~\cite{salimans2022progressive} is used by reparameterizing the denoiser as  
$\rvepsilon_\theta(\rvd_t, \rvx, t) = \frac{1}{\alpha_t} (\rvd_t - \sigma_t \rvv_\theta(\rvd_t, \rvx, t))$.
During inference, $\rvd_0$ is obtained by starting from a Gaussian noise $\rvd_T \sim \mathcal{N}(0, I)$, and iteratively applying the denoiser $\rvepsilon_\theta(\cdot)$. Finally, the decoder $\gD$ reconstructs the depth map $\hat{\rmD}$ from the estimated clean latent $\hat{\rvd}_0$.
\vspace{-1mm}
\subsection{Controllable Video Generation Models}
\vspace{-1mm}
\label{sec:preliminary_video}
Using image-to-video (I2V) diffusion models~\cite{2023i2vgenxl,blattmann2023stable}, recent video generation methods synthesize videos from a single image by modeling the conditional distribution $p(\rmX^{(1:F)}~|~\rmX^{(1)})$, where $\rmX^{(1)} \in \mathbb{R}^{3 \times H \times W}$ is an input image and $\rmX^{(1:F)} \in \mathbb{R}^{F \times 3 \times H \times W}$ represents the generated video frames.
Controllable video generation methods~\cite{lin2024ctrl} incorporate additional guidance (e.g., depth or normal) to enforce structural consistency. This extends the formulation to $p(\rmX^{(1:F)}~|~\rmX^{(1)}, \rmD^{(1:F)})$, where $\rmD^{(1:F)} \in \mathbb{R}^{F \times H \times W}$ represents depth or normal maps that provide geometric information. 
In this work, we formulate video depth and normal estimation as an image-to-video generation task, building our system upon both image geometry estimation models~\cite{marigold} and conditional video generation frameworks~\cite{lin2024ctrl}.

\vspace{-1.mm}
\section{GeoMan}
\vspace{-1mm}
GeoMan aims to estimate temporally consistent and accurate depth and normal maps from monocular human videos. 
Our key idea is to frame video geometry estimation as an image-to-video generation task (Sec.~\ref{sec:image-to-video}), leveraging the pre-trained priors of image-to-video models learned with large-scale video datasets.
To preserve human scale information in the estimated depth maps, we propose a novel human-centered metric depth representation in Sec.~\ref{sec:metric-depth} that computes root-relative distances as the depth measurement.
We explain the process of synthesizing training data to train our models using 3D and 4D human scans in Sec.~\ref{sec:dataset}.
\vspace{-1mm}

\subsection{Temporal Human Geometry Estimation}  
\vspace{-1mm}
\label{sec:image-to-video}
Given a monocular RGB video of a single moving human,  $\rmX^{(1:F)} \in \mathbb{R}^{F \times 3 \times H \times W}$, our objective is to estimate dynamic human geometry, including depth maps $\rmD^{(1:F)} \in \mathbb{R}^{F \times 3 \times H \times W}$ and normal maps $\rmN^{(1:F)} \in \mathbb{R}^{F \times 3 \times H \times W}$ for all $F$ frames in the video. 
To reduce dependency on 4D training data, we decompose this complex task into image-geometry estimation and image-to-video synthesis. 
We begin by training an Image-to-Geometry Estimation Model (I2G) that estimates the depth and normal map of the first frame in a video. 
This initial estimation serves as guidance for our Video-to-Geometry Estimation Model (V2G), which predicts the depth or normal maps for subsequent frames. 
As demonstrated in Tab.~\ref{tab:ablation}(a) and Fig.~\ref{fig:naive-vs-geoman}, our approach reduces reliance on extensive 4D training datasets, while achieving superior performance and strong generalization ability.
Fig.~\ref{fig:overview}(a) shows an overview of GeoMan.

\begin{figure}[!t]
    \centering
    \includegraphics[width=\columnwidth]{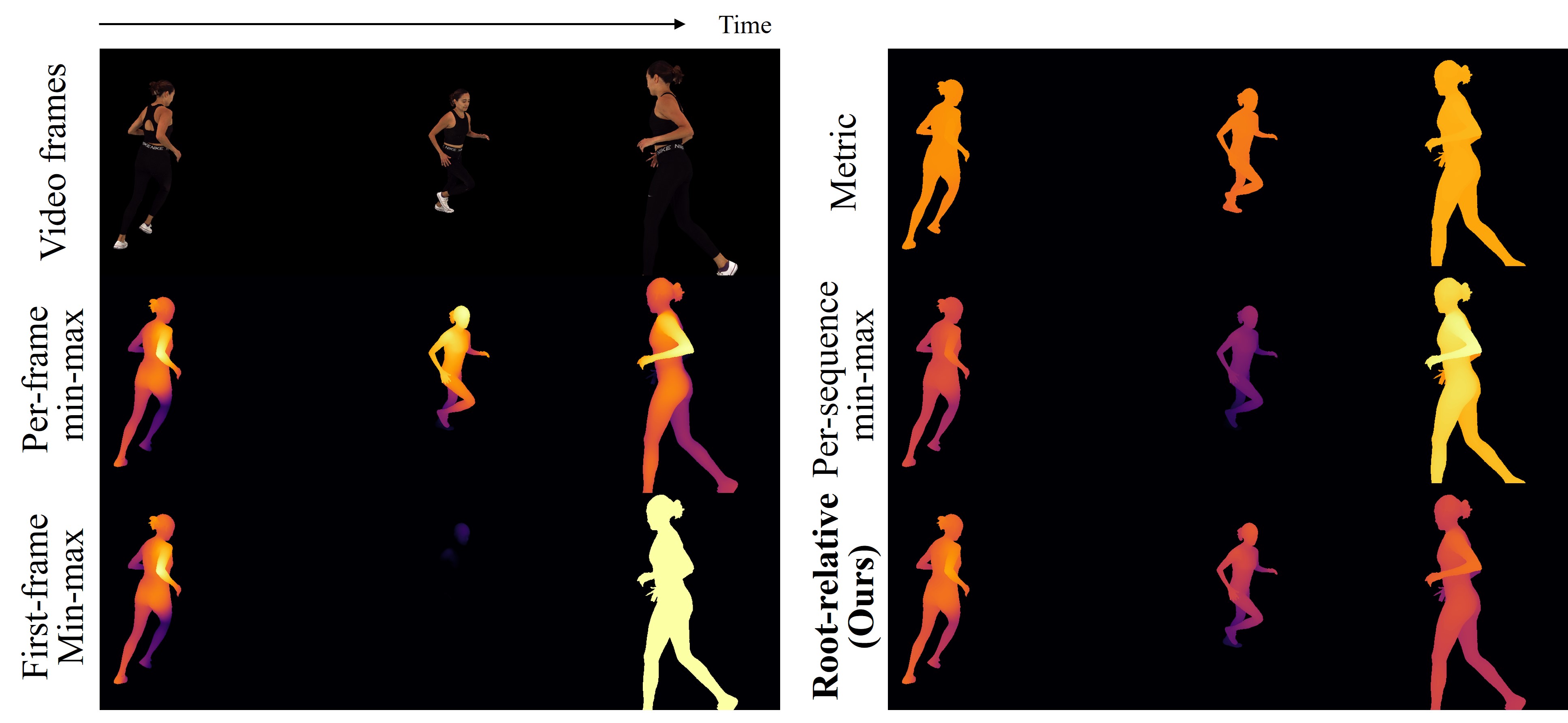}
    \vspace{-2.2em}
    \caption{Comparison of depth representations: Our representation offers the highest fidelity and improves temporal modeling.}
    \label{fig:depth-comparison}
    \vspace{-0.5em}
\end{figure}

\begin{table}[!t]
\begin{adjustbox}{width=\linewidth}
\vspace{-0.8em}
\centering
\begin{tabular}{cccc}
\hline
\textbf{Method} & \textbf{Aff-invariant} & \textbf{Metric}  & \textbf{Root-relative} \\
\hline
Formulation & $\frac{D - \min(D)}{\max(D) - \min(D)}$ & $D$ & $D - d_{\text{root}}$ \\
Range & [0, 1] & [0, $\infty$] & [$-h/2$, $h/2$] \\
Scale preservation & \xmark & \cmark & \cmark \\
Temporal consistency & \xmark & \cmark & \cmark \\
Detail preservation & \cmark & \xmark & \cmark \\
\hline
\end{tabular}
\end{adjustbox}
\vspace{-0.8em}
\caption{Comparison of different depth representations. $h$ is the maximum human height. We divide $h$ by 2 because the root keypoint is roughly in the middle of the body.}
\label{tab:depth_representation}
\vspace{-1.5em}
\end{table}

\paragraph{Image-to-Geometry Estimation Model (I2G).}
We predict geometry for the first frame by modeling the conditional distributions $p(\rmD^{(1)}~|~\rmX^{(1)})$ and $p(\rmN^{(1)}~|~\rmX^{(1)})$. 
Inspired by Marigold~\cite{marigold}, we learn a LDM~\cite{han2023high} within the latent space of a pre-trained VAE model~\cite{kingma2013auto} for depth and normal estimation. 
More details can be found in Sec.~\ref{sec:preliminary_image}, Sec. B.1 of the supplementary material and in~\cite{marigold}.
Note that we learn two separate diffusion models for depth and normal prediction, which yields better performance compared to sharing a single model across two modalities.

\paragraph{Video-to-Geometry Estimation Model (V2G).}
Given the geometry estimations from the first frame as discussed above, we formulate geometry estimation for the subsequent frames as modeling the conditional distributions $p(\rmD^{(1:F)}~|~\rmX^{(1:F)}, \rmD^{(1)})$ and $p(\rmN^{(1:F)}~|~\rmX^{(1:F)}, \rmN^{(1)})$. 
%
We adopt a ControlNet-based diffusion model~\cite{lin2024ctrl} that takes the video frames $\rmX^{(1:F)}$ as control inputs and the first frame's geometry (depth $\rmD^{(1)}$ or normal $\rmN^{(1)}$) as a reference image. 
As illustrated in Fig.~\ref{fig:overview}(a), we learn a unified model for both depth and normal estimation, which can be achieved by simply switching the reference image input between depth $\rmD^{(1)}$ or normal map $\rmN^{(1)}$ of the first frame.
Our intuition is that by framing video geometry estimation as an image-to-video synthesis task, depth and normal maps of the first frame are solely used as a reference image while the fundamental task (i.e., image-to-video synthesis) remains the same for both modalities, hence it enables us to share a single model across two modalities. 
In the following, we assume the reference image is a normal map, depth estimation can be similarly performed by changing the reference image to a depth map. 

At each training step, we map the normal maps to the latent space by a pre-trained encoder and add noise to the normal latents $\rvn^{(1:F)}$ at a random timestep $t$. We concatenate the latents of the first frame's normal with noisy normal latents and pass them to the video denoising model to estimate the added noise, where the input video frames are injected into the denoising model as control signal using a ControlNet. 
We adopt the $v$-objective~\cite{salimans2022progressive} to ensure stable performance.  
We optimize the denoiser by minimizing the difference between the estimated and ground truth velocity:  
\vspace{-2mm}
\begin{equation}
    \mathcal{L}_{\text{V2G}} = \mathbb{E}_{\rvn^{(1:F)}_0, \rvepsilon, t} \left\| \rvv_t - \rvv_\phi(\rvn^{(1:F)}_t, \rvn^{(1)}, \rmX^{(1:F)}, t) \right\|^2_2,
\label{eq:diffusion_objective_dyn}
\vspace{-2mm}
\end{equation}  
where $\rvv_\phi$ denotes our velocity prediction model, $\rvv_t = \alpha_t \rvepsilon - \sigma_t \rvn^{(1:F)}_0$,  
$\rvepsilon \sim \mathcal{N}(0, I)$, $t \sim \mathcal{U}(T)$, and $\rvn^{(1:F)}_t$ represents the noisy normal latents at timestep $t$. 
During inference, we start from a sequence of random noise maps and gradually denoise them to normal maps, using the video frames as control signals and the first frame's normal prediction from I2G as the reference image.

Compared to existing video depth and normal estimation works~\cite{xiu2022econ,khirodkar2024sapiens,Jafarian_2021_CVPR_TikTok}, our V2G model provides two key advantages:
1) \textit{Efficient utilization of pre-trained representations}: By leveraging a pre-trained image-to-video model, we exploit the powerful prior learned from large-scale video datasets.
This enables efficient training with less data while enhancing both temporal consistency and geometric accuracy.
2) \textit{Multi-modality compatibility}: The model utilizes shared weights across tasks, toggling between depth and normal estimation by switching the reference image. This enables simultaneous training for both modalities, improving prediction accuracy (See Table~\ref{tab:ablation}). 
As as result, our V2G model demonstrates strong generalizability, effectively handling \textit{long video} sequences and \textit{multi-person} scenarios. Please refer to Sec. C.1 in supplementary for details.

\begin{figure*}[!t]
    \centering
    \includegraphics[width=0.9\textwidth]{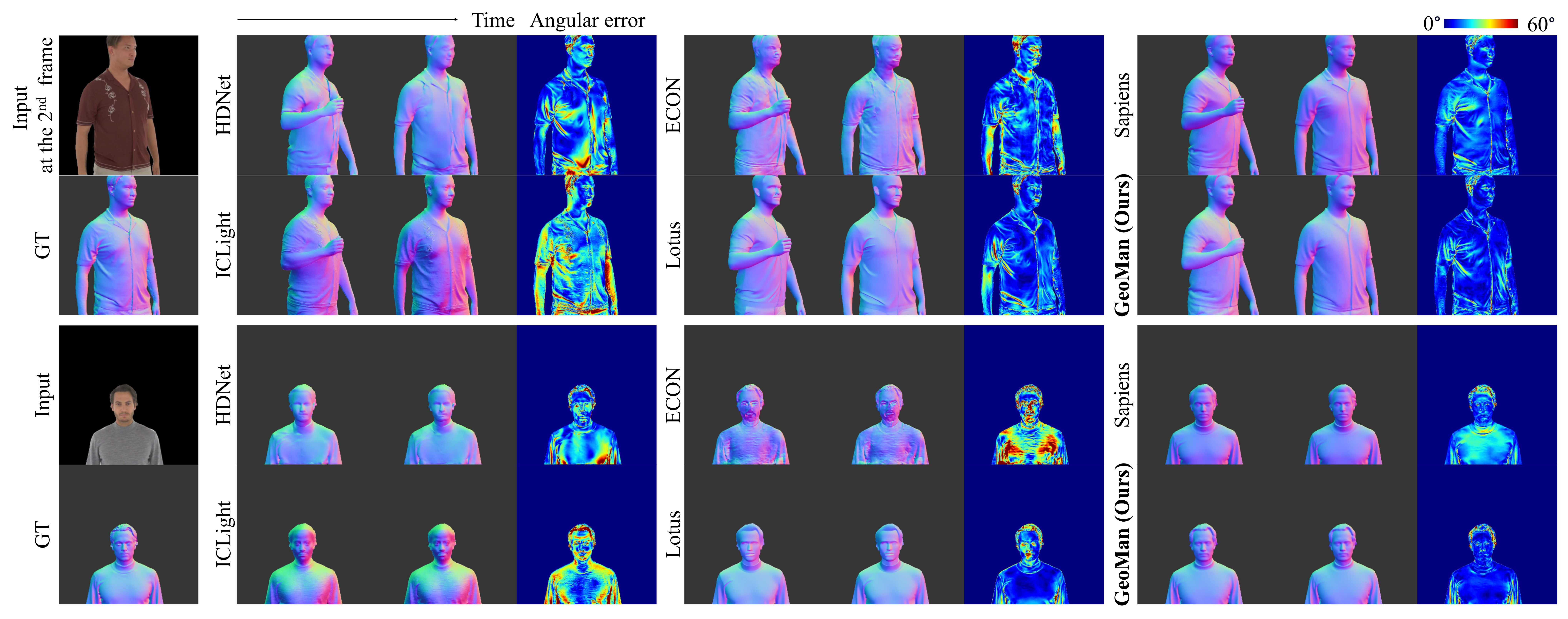}
    \vspace{-1em}
    \caption{Zero-shot normal estimation comparison on ActorsHQ. \textbf{Left}: Predicted normal for the first frame. \textbf{Middle}: Predicted normal for the second frame. \textbf{Right}: Angular error visualization for the second frame.}
    \label{fig:normal}
\end{figure*}
\begin{table*}[!t]\centering

\scriptsize
\vspace{-1em}
\centering
\begin{adjustbox}{width=0.9\linewidth}
\begin{tabular}{lcccccccccccc}\toprule
&\multicolumn{4}{c}{\textbf{Performance on moving subject videos}} &\multicolumn{4}{c}{\textbf{Performance on moving camera videos}} &\multicolumn{3}{c}{\textbf{Temporal consistency}} \\\cmidrule{2-12}
&Mean↓ &Median↓ &11.25°↑ &30°↑ &Mean↓ &Median↓ &11.25°↑ &30°↑ &OPW↓ &TC-Mean↓ &TC-11.25°↑ \\\midrule
ECON~\cite{xiu2022econ} &29.228 &23.766 &17.963 &62.464 &27.202 &22.089 &19.652 &66.682 &0.173 &16.371 &50.469 \\
ICLight~\cite{iclight} &27.996 &23.555 &16.384 &64.276 &26.975 &22.529 &17.595 &66.567 &0.122 &11.476 &66.866 \\
HDNet~\cite{Jafarian_2021_CVPR_TikTok}  &24.477 &20.329 &21.737 &71.527 &23.954 &19.867 &22.493 &72.764 &0.134 &12.398 &62.957 \\
GeoWizard~\cite{fu2024geowizard} &22.714 &19.151 &23.446 &75.610 &22.054 &18.520 &24.796 &76.981 &0.178 &16.584 &43.846 \\
Lotus~\cite{he2024lotus} &18.780 &14.937 &34.947 &83.743 &18.101 &14.444 &36.605 &85.217 &\ul{0.079} &8.782 &79.397 \\
Sapiens*~\cite{khirodkar2024sapiens} &18.338 &14.526 &36.498 &85.009 &18.048 &14.425 &36.706 &85.760 &0.094 &9.494 &77.567 \\
Sapiens~\cite{khirodkar2024sapiens} &\ul{16.278} &\ul{13.015} &\ul{41.245} &\textbf{89.177} &\ul{16.099} &\ul{12.941} &\ul{41.539} &\textbf{89.617} &0.082 &\ul{7.861} &\ul{82.803} \\
\textbf{GeoMan (Ours)} &\textbf{16.185} &\textbf{12.334} &\textbf{45.217} &\ul{87.784} &\textbf{15.535} &\textbf{11.657} &\textbf{48.168} &\ul{88.789} &\textbf{0.070} &\textbf{6.876} &\textbf{96.757} \\
\bottomrule
\end{tabular}
\end{adjustbox}
\vspace{-1em}
\caption{Comparison on zero-shot normal estimation on ActorsHQ dataset. Sapiens*: fine-tuned on our dataset. GeoMan outperforms state-of-the-art methods despite using only public data, unlike Sapiens which was trained using large-scale proprietary data. The drop in performance of Sapiens when fine-tuned on our dataset highlights the importance of its proprietary data in its performance.}
\label{tab:normal_sota}
\vspace{-.5em}
\end{table*}

\vspace{-1mm}
\subsection{Human-Centered Root-Relative Depth}
\vspace{-1mm}
\label{sec:metric-depth}
A major challenge in monocular depth estimation lies in selecting an optimal depth representation. 
Most existing methods~\cite{khirodkar2024sapiens,marigold} employ an affine-invariant depth representation with per-frame normalization, which loses crucial information about human scale, and leads to inferior temporal stability. 
Recent video depth estimation methods~\cite{khirodkar2024sapiens,marigold} introduce per-sequence normalization to improve stability across sequences, though they often struggle to capture fine details effectively.
Conversely, methods~\cite{bhat2023zoedepth,bochkovskii2024depthpro} predicting metric depth directly often sacrifice spatial details, compromising the accuracy of geometry estimation.
Fig.~\ref{fig:depth-comparison} provides an intuitive comparison of different depth representations.

To address these limitations, we propose a human-centered root-relative depth representation that preserves metric-scale details while only discarding global translation. This representation ensures accurate human size and scale, which is crucial for tasks requiring precise human geometry (e.g., human reconstruction).
As shown in Fig.~\ref{fig:overview}(b), the root-relative depth of each pixel is computed as the distance between the human root (i.e., the pelvis joint) and each point on the human surface.
Formally, the root-relative depth $\rmD^{\text{root-rel}}$ is computed as $\rmD^{\text{root-rel}} = \rmD^{\text{metric}} - d^{\text{root}}$, where $\rmD^{\text{metric}}$ is the metric depth, $d^{\text{root}}$ is the depth of the root joint, which can be obtained via pose estimation techniques~\cite{wang2024tram,goel2023humans}. 
For training, we obtain the root joint's depth using the SMPL-X~\cite{SMPL-X:2019} annotation in the datasets. Given the SMPL-X pelvis joint location in world space  \(\mathbf{P}_w \in \mathbb{R}^3\), the camera rotation matrix \(\mathbf{R} \in \mathbb{R}^{3 \times 3}\) and translation vector \(\mathbf{T} \in \mathbb{R}^{3}\), we compute the root depth as $d^{\text{root}} = (\mathbf{R} \mathbf{P}_w + \mathbf{T})_z$, as illustrated in Fig.~\ref{fig:overview}(b).

We compare depth representations in Table~\ref{tab:depth_representation}. Root-relative depth offers several key advantages:  
1) \textit{Human scale preservation:} It retains the relative depth information of the person in metric scale, allowing inference of real human size. 
In contrast, affine-invariant depth always normalizes this size to 1, discarding meaningful scale information.  Additionally, absolute metric depth can be easily predicted by combining root-relative depth with root depth estimated by advanced human pose prediction models~\cite{wang2024tram,goel2023humans}.
2) \textit{Temporal consistency:} Root-relative depth is more temporally stable, as it avoids per-frame normalization using varying $\min(D)$ and $\max(D)$, which is required for affine-invariant depth and can introduce temporal artifacts.  
3) \textit{Geometric detail preservation:} The range of root-relative depth is constrained to $[-h/2, h/2]$, determined by the maximum human height ($h$) and independent of global position. In contrast, metric depth spans a much wider range (e.g., $[0, \infty]$), making it harder for networks to focus on fine-grained geometry, as they must learn large global variations.

\subsection{Training with Limited Geometric Video Data}
\vspace{-1mm}
\label{sec:dataset}
We synthesize our training data by rendering 3D scans from the THuman-2.0~\cite{yu2021function4d} dataset and 4D scans from the XHumans~\cite{shen2023xavatar,wang20244ddress} dataset using diverse camera distance, pitch angles, and focal lengths.
To train the I2G model, we rendered 40K frames from the THuman-2.0~\cite{yu2021function4d} dataset, which consists of 525 high-resolution, textured 3D scans with approximately 300K vertices per scan.
For training the V2G model, we addressed the limited diversity in sequences and subjects by combining data from the XHumans~\cite{shen2023xavatar} dataset (233 sequences across 20 subjects) with augmented data derived from THuman-2.0. To leverage the static 3D scans in THuman-2.0, we transform multiview images into simulated videos that mimic camera rotation.

In total, our model is trained using only 20 subjects in 4D~\cite{shen2023xavatar} and around 500 subjects in 3D~\cite{yu2021function4d} from public datasets. This dataset is significantly smaller compared to the large-scale proprietary datasets employed by state-of-the-art methods~\cite{khirodkar2024sapiens, Jafarian_2021_CVPR_TikTok}. Despite limited data, our approach demonstrates strong performance, underscoring its efficiency and robustness.
\vspace{-1mm}
\section{Experiments}
\vspace{-2mm}
We present comprehensive videos at \href{https://research.nvidia.com/labs/dair/geoman}{our project page} to visually evaluate dynamic geometry estimation. 
The supplementary material also includes implementation details, evaluation specifics, additional results on the Goliath dataset~\cite{martinez2024codec}, in-the-wild videos, long videos, and multi-person scenarios.

\noindent \textbf{Evaluation Dataset.}
We use ActorsHQ~\cite{isik2023humanrf} as our primary evaluation dataset for its high-quality dynamic geometry ($\sim$30K vertices) and independence from training sets. 
For evaluation, we generate 192 videos of moving subjects with static cameras and 256 videos with moving cameras and static subjects. The full evaluation dataset includes more than 10K frames rendered at 512$\times$512 resolution.
Additional results on the Tiktok~\cite{Jafarian_2022_TPAMI} and the Goliath~\cite{martinez2024codec} dataset can be found in Sec. C.2 and Sec. C.3 of the supplementary.

\begin{figure*}[!t]
    \centering
    \includegraphics[width=0.9\textwidth]{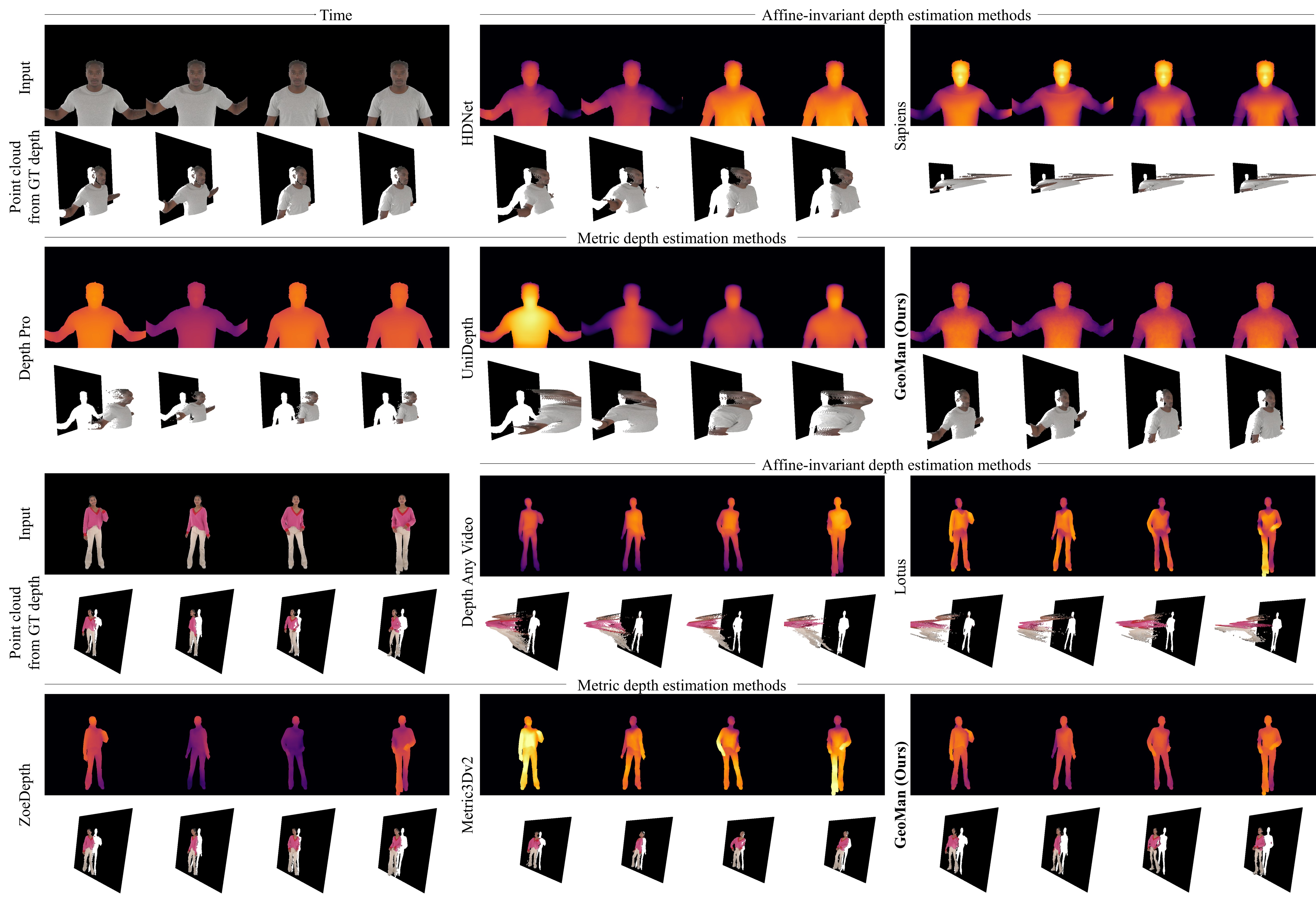}
    \vspace{-1em}
    \caption{
    Comparison with existing depth estimation models.
    GeoMan achieves state-of-the-art performance in both depth prediction (top row in each result) and point cloud reconstruction (bottom row), excelling in temporal consistency,  fidelity, and scale preservation.
    }
    \vspace{-1.em}
    \label{fig:depth_sota_comparison}
\end{figure*}

\begin{table*}[!t]\centering

\scriptsize
\label{tab:quanti}
\centering
\begin{adjustbox}{width=0.95\linewidth}
\begin{tabular}{lcccccccccccc}\toprule
&\multicolumn{4}{c}{\textbf{Performance on moving subject videos}} &\multicolumn{4}{c}{\textbf{Performance on moving camera videos}} &\multicolumn{3}{c}{\multirow{2}{*}{\textbf{Temporal consistency}}} \\\cmidrule{2-9}
&\multicolumn{2}{c}{Absolute} &\multicolumn{2}{c}{Optimizing scale + shift} &\multicolumn{2}{c}{Absolute} &\multicolumn{2}{c}{Optimizing scale + shift} & & & \\\cmidrule{2-12}
&$\delta$1↑ &SI-log10↓ &AbsRel↓ &$\delta<$1.05↑ &$\delta$1↑ &SI-log10↓ &AbsRel↓ &$\delta<$1.05↑ &OPW↓ &TC-RMSE↓ &TC-$\delta$1↑ \\\midrule
ZoeDepth~\cite{bhat2023zoedepth} &0.016 &2.710 &0.016 &0.960 &0.000 &2.487 &0.018 &0.955 &{0.058} &{0.017} &{0.830} \\
UniDepth~\cite{piccinelli2024unidepth} &0.058 &8.027 &0.025 &0.883 &0.003 &7.790 &0.022 &0.915 &0.128 &0.031 &0.564 \\
Depth pro~\cite{bochkovskii2024depthpro} &0.127 &5.297 &0.026 &0.865 &0.103 &5.804 &0.025 &0.882 &0.201 &0.046 &0.388 \\
Metric3Dv2~\cite{hu2024metric3d} &0.491 &6.213 &0.016 &0.958 &0.697 &5.804 &0.019 &0.944 &0.085 &0.023 &0.743 \\
\textbf{GeoMan (Ours)} &\ul{0.972} &\textbf{2.116} &\textbf{0.011} &\textbf{0.984} &\ul{0.987} &\textbf{1.775} &\textbf{0.009} &\textbf{0.992} &\ul{0.038} &\textbf{0.015} &\ul{0.898} \\
\midrule
\textbf{GeoMan w GT root-depth (Ours)} &\textbf{0.998} &\ul{2.118} &\textbf{0.011} &\textbf{0.984} &\textbf{0.998} &\textbf{1.775} &\textbf{0.009} &\textbf{0.992} &\textbf{0.032} &\textbf{0.015} &\textbf{0.915} \\
\bottomrule
\end{tabular}
\end{adjustbox}
\vspace{-1em}
\caption{Comparison with metric depth estimation methods on zero-shot settings on ActorsHQ dataset.}\
\label{tab:metric_depth_sota}
\end{table*}

\begin{table*}[!t]\centering

\scriptsize
\vspace{-1em}
\label{tab:quanti}
\centering
\begin{adjustbox}{width=0.95\linewidth}
\begin{tabular}{lcccccccccccc}\toprule
&\multicolumn{4}{c}{\textbf{Performance on moving subject videos}} &\multicolumn{4}{c}{\textbf{Performance on moving camera videos}} &\multicolumn{3}{c}{\multirow{2}{*}{\textbf{Temporal consistency}}} \\\cmidrule{2-9}
&\multicolumn{2}{c}{Optimizing shift} &\multicolumn{2}{c}{Optimizing scale + shift} &\multicolumn{2}{c}{Optimizing shift} &\multicolumn{2}{c}{Optimizing scale + shift} & & & \\\cmidrule{2-12}
&AbsRel↓ &$\delta<$1.05↑ &AbsRel↓ &$\delta<$1.05↑ &AbsRel↓ &$\delta<$1.05↑ &AbsRel↓ &$\delta<$1.05↑ &OPW↓ &TC-RMSE↓ &TC-$\delta$1↑ \\\midrule
HDNet~\cite{Jafarian_2021_CVPR_TikTok} &0.031 &0.808 &0.024 &0.877 &0.030 &0.821 &0.021 &0.921 &0.108 &0.027 &0.604 \\
Sapiens~\cite{khirodkar2024sapiens} &0.040 &0.680 &\textbf{0.009} &\textbf{0.986} &0.041 &0.663 &\ul{0.010} &\ul{0.987} &0.058 &0.018 &0.792 \\
Sapiens*~\cite{khirodkar2024sapiens} &0.021 &0.915 &0.015 &0.959 &0.017 &0.960 &0.013 &0.980 &0.054 &0.018 &0.816 \\
Marigold~\cite{marigold} &0.021 &0.901 &0.021 &0.901 &0.015 &0.956 &0.017 &0.951 &0.074 &0.027 &0.547 \\
GeoWizard~\cite{fu2024geowizard} &0.018 &0.935 &0.019 &0.930 &0.014 &0.967 &0.017 &0.952 &0.053 &0.023 &0.667 \\
Lotus~\cite{he2024lotus} &0.022 &0.899 &0.019 &0.919 &0.016 &0.954 &0.015 &0.962 &0.077 &0.025 &0.637 \\
DepthCrafter~\cite{hu2024-DepthCrafter} &\ul{0.016} &\ul{0.944} &0.013 &0.967 &\ul{0.012} &\ul{0.976} &0.018 &0.945 &\ul{0.037} &\ul{0.016} &\ul{0.914} \\
Depth-Any-Video~\cite{yang2024depthanyvideo} &0.018 &0.928 &0.015 &0.955 &0.014 &0.971 &0.018 &0.948 &0.039 &0.017 &0.823 \\
\textbf{GeoMan (Ours)} &\textbf{0.012} &\textbf{0.978} &\ul{0.011} &\ul{0.984} &\textbf{0.009} &\textbf{0.991} &\textbf{0.009} &\textbf{0.992} &\textbf{0.032} &\textbf{0.014} &\textbf{0.915} \\
\bottomrule
\end{tabular}
\end{adjustbox}
\vspace{-1em}
\caption{Comparison with affine-invariant depth estimation methods in zero-shot on ActorsHQ. Sapiens*: fine-tuned on our dataset.}\
\label{tab:affine_depth_sota}
\vspace{-2em}
\end{table*}

\subsection{Surface Normal Estimation}
\vspace{-1mm}
\noindent \textbf{Baselines.} 
We compare our method against baselines including HDNet~\cite{Jafarian_2021_CVPR_TikTok,Jafarian_2022_TPAMI}, ECON~\cite{xiu2022econ}, and Sapiens~\cite{khirodkar2024sapiens}, which are state-of-the-art normal estimation models trained on human-centric data. 
Notably, methods like Sapiens and HDNet were trained using proprietary human-centric datasets, whereas our model was trained on publicly available datasets.
Additionally, we compare against models trained for generic scenes, such as GeoWizard~\cite{fu2024geowizard}, Lotus~\cite{he2024lotus}, and ICLight~\cite{iclight}. For all comparisons, we used the official checkpoints provided by the  authors. Additionally, we report the performance of Sapiens, the best-performing baseline, fine-tuned on our dataset.

\noindent \textbf{Metrics.}
We report mean and median angular error, along with the percentage of pixels within $t^\circ$ error for $t \in \{11.25^\circ, 30^\circ\}$. To assess temporal consistency, we provide the optical flow-based warping metric (OPW)~\cite{brox2004high}, as well as the optical flow-based angular error (TC-Mean) and the $11.25^\circ$ metric (TC-$11.25^\circ$).

\noindent \textbf{Results.} 
As shown in Table~\ref{tab:normal_sota}, GeoMan outperforms previous approaches, achieving higher accuracy in normal prediction and better temporal consistency. While GeoMan significantly outperforms other methods, Sapiens~\cite{khirodkar2024sapiens} performs competitively. However, it is important to note that Sapiens was trained on a proprietary dataset. For a fair comparison, we also fine-tune Sapiens's pre-trained MAE model on our dataset. As we can see, under this more fair setting, GeoMan outperforms Sapiens by a notable margin. Fig.~\ref{fig:normal} presents a qualitative comparison, demonstrating that GeoMan delivers more consistent estimations over time and higher fidelity.%

\begin{table*}[!t]\centering

\scriptsize
\label{tab:quanti}
\centering
\begin{adjustbox}{width=0.9\linewidth}
\begin{tabular}{lcccccccccccc}\toprule
&\multicolumn{5}{c}{\textbf{Normal}} &\multicolumn{6}{c}{\textbf{Depth}} \\\cmidrule{2-12}
&\multicolumn{2}{c}{Angular difference} &\%within t° &\multicolumn{2}{c}{Temporal consistency} &\multicolumn{2}{c}{Optimizing shift} &\multicolumn{2}{c}{Optimizing scale + shift} &\multicolumn{2}{c}{Temporal consistency} \\\cmidrule{2-12}
&Mean↓ &Median↓ &11.25°↑ &OPW↓ &TC-Mean↓ &AbsRel↓ &$\delta<$1.05↑ &AbsRel↓ &$\delta<$1.05↑ &OPW↓ &TC-RMSE↓ \\\midrule
&\multicolumn{11}{c}{\textbf{(a) Naive extension vs GeoMan across training steps}} \\
Naïve-5K-steps &25.305 &22.732 &15.602 &0.076 &7.458 &0.038 &0.720 &0.028 &0.856 &0.049 &0.019 \\
Naïve-30K steps &19.669 &16.628 &28.405 &0.072 &7.169 &0.020 &0.926 &0.017 &0.943 &0.040 &0.016 \\\arrayrulecolor{black!30}\midrule
\textbf{GeoMan-5K steps} &17.499 &13.541 &40.340 &0.069 &6.781 &0.014 &0.964 &0.013 &0.970 &0.036 &\textbf{0.014} \\
\textbf{GeoMan-30K steps} &\textbf{16.185} &\textbf{12.334} &\textbf{45.217} &0.070 &6.876 &\textbf{0.012} &\textbf{0.978} &\textbf{0.011} &\textbf{0.984} &\textbf{0.032} &\textbf{0.014} \\\arrayrulecolor{black}\midrule
&\multicolumn{11}{c}{\textbf{(b) Naïve extension vs GeoMan on 1/10 training data (at 30K steps)}} \\
Naïve w/ 1/10 data &22.329 &19.626 &20.568 &0.068 &6.833 &NaN &NaN &NaN &NaN &NaN &NaN \\
\textbf{GeoMan w/ 1/10 data} &\textbf{17.679} &\textbf{13.559} &\textbf{40.263} &\textbf{0.065} &\textbf{6.451} &\textbf{0.013} &\textbf{0.975} &\textbf{0.011} &\textbf{0.982} &\textbf{0.032} &\textbf{0.014} \\
\midrule
&\multicolumn{11}{c}{\textbf{(c) Multimodality of V2G}} \\
Unimoal &16.260 &12.361 &45.091 &0.073 &7.190 &\textbf{0.012} &0.976 &\textbf{0.011} &0.983 &0.038 & {0.015} \\
\textbf{Multimodal} &\textbf{16.185} &\textbf{12.334} &\textbf{45.217} &\textbf{0.070} &\textbf{6.876} &\textbf{0.012} &\textbf{0.978} &\textbf{0.011} &\textbf{0.984} &\textbf{0.032} &\textbf{0.014} \\\midrule
&\multicolumn{11}{c}{\textbf{(d) Source of the first frame for V2G}} \\
GT+V2G &\textbf{14.028} &\textbf{10.101} &\textbf{54.924} &0.078 &7.536 &\textbf{0.009} &\textbf{0.985} &\textbf{0.009} &\textbf{0.988} &\textbf{0.032} &\textbf{0.014} \\
\textbf{I2G+V2G} &16.185 &12.334 &45.217 &\textbf{0.070} &\textbf{6.876} &0.012 &0.978 &0.011 &0.984 &\textbf{0.032} &\textbf{0.014} \\
\bottomrule
\end{tabular}
\end{adjustbox}
\vspace{-1em}
\caption{Ablation studies. We validate the impact of various design choices.}
\vspace{-2em}
\label{tab:ablation}
\end{table*}

\vspace{-1mm}
\subsection{Depth Estimation}
\vspace{-1mm}
\noindent \textbf{Baselines.} 
For metric depth estimation, we compare with state-of-the-art methods including Metric3Dv2~\cite{hu2024metric3d}, UniDepth~\cite{piccinelli2024unidepth}, ZoeDepth~\cite{bhat2023zoedepth}, and Depth Pro~\cite{bochkovskii2024depthpro}. 
For affine-invariant depth estimation, we benchmark against image-based models such as Sapiens~\cite{khirodkar2024sapiens}, Marigold~\cite{marigold}, GeoWizard~\cite{fu2024geowizard}, and Lotus~\cite{he2024lotus}. 
Additionally, we include video-based methods like DepthCrafter~\cite{hu2024-DepthCrafter} and Depth Any Video~\cite{yang2024depthanyvideo}, which are trained on large-scale video datasets. For all baselines, we used official checkpoints.

\noindent \textbf{Metrics.}
For metric depth estimation, we follow existing methods~\cite{bochkovskii2024depthpro} and report $\delta_1$ and SI-log10 (Scale-Invariant Logarithmic Error). The root depth is obtained using 3D pose estimation method~\cite{wang2024tram} to convert our predictions to metric depth. For evaluating affine-invariant depth estimation methods, we align the predicted depth maps with the ground truth using two approaches: (1) optimizing only the shift to highlight the benefits of GeoMan's learned human scale, and (2) optimizing both scale and shift before computing the metrics. We report AbsRel (absolute relative error) and $\delta < 1.05$ (the percentage of pixels where $\max(d / \hat{d}, \hat{d} / d) < 1.05$). We further report affine-invariant metrics for metric depth estimation methods to provide a comprehensive assessment. 
To evaluate temporal consistency, we report the optical flow-based warping metric (OPW)~\cite{wang2022less},  RMSE, and the $\delta_1$ metric. Details of all metrics can be found in Sec. B.3 in the supplementary.

\noindent \textbf{Results.} 
Table~\ref{tab:metric_depth_sota} summarizes the results for metric depth estimation. GeoMan significantly outperforms other methods, demonstrating its effectiveness. Notably, we outperform the second-best method Metric3Dv2 by ~100\% and ~65\% in terms of $\delta_1$ and SI-log10, respectively. ZoeDepth performs best in terms of temporal consistency but yields significantly worse accuracy metrics, due to overly smooth predictions (See Fig.~\ref{fig:depth_sota_comparison}).  
Additionally, instead of using the pose estimation method VIMO~\cite{wang2024tram} to estimate the root joint depth, we also evaluate performance using ground-truth root depth. This further improves the results, highlighting the potential for additional enhancements in future works.
Table~\ref{tab:affine_depth_sota} compares GeoMan with affine-invariant depth estimation methods. When optimizing only the shift, our model achieves the best results, showcasing the advantages of our proposed root-relative depth representation in capturing human-scale information. Our method outperforms most baselines with both scale and shift optimization while maintaining superior temporal consistency. 

Fig.~\ref{fig:depth_sota_comparison} presents a qualitative comparison with baselines. 
All depth maps are renormalized using sequence-wise min-max scaling within the human mask to ensure consistency, following~\cite{khirodkar2024sapiens}. Compared to per-frame estimation models, our approach produces more temporally stable results. While UniDepth~\cite{piccinelli2024unidepth} and DepthCrafter~\cite{hu2024-DepthCrafter} improve temporal consistency, our model achieves higher accuracy both visually and quantitatively.
The second row of Fig.~\ref{fig:depth_sota_comparison} compares point cloud reconstructions (PCD) generated by our method and baselines, where pixels are lifted into 3D space using the predicted depth and camera parameters provided in the dataset. Prior methods produce distorted and temporally inconsistent reconstructions, whereas GeoMan ensures accurate scaling and superior temporal consistency.

\subsection{Ablation Studies and Analysis}  
\vspace{-1mm}
\begin{figure}[!t]
    \centering
    \includegraphics[width=\columnwidth]{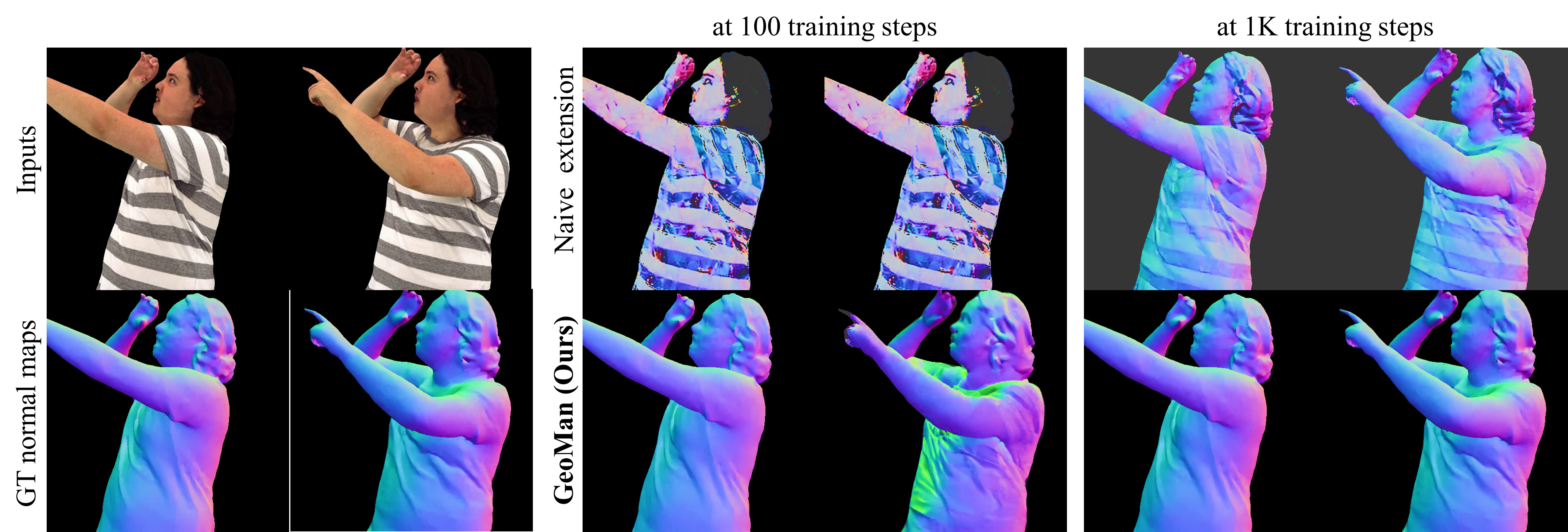}
    \vspace{-2em}
    \caption{Comparison of naive extension and GeoMan across training steps. Results here are evaluated on the validation data.}%
    \vspace{-2em}
    \label{fig:naive-vs-geoman}
\end{figure}

\noindent \textbf{Naive Extension vs. GeoMan.}  
We compare our approach with a naïve extension of \cite{marigold, hu2024-DepthCrafter}. In this extension, instead of decomposing the task into I2G and V2G as in GeoMan, a video diffusion model is trained to predict normal or depth maps for all frames in a single pass without employing a control module. This design suffers from the task gap and fails to effectively utilize prior knowledge, resulting in suboptimal performance.
Evaluation results at different training iterations are presented in Table~\ref{tab:ablation}(a). Notably, GeoMan achieves superior performance even at early ($\sim$5K) iterations, outperforming the naïve extension approach at 30K iterations. This performance gap arises because directly predicting geometric information for all frames deviates significantly from the original task of the video diffusion model. As a result, naïve extension fails to effectively leverage its prior, as can be seen in Fig.~\ref{fig:naive-vs-geoman}.
We further demonstrate GeoMan's superior generalization ability by training both models on a reduced dataset (1/10 of the original), the naïve extension approach suffers significantly, with poor performance in the normal model and NaN scores in the depth model as presented in Table~\ref{tab:ablation}(b). 
This highlights the effectiveness of our I2G and V2G model in utilizing pre-trained priors for its representation and generalization capacity.

\noindent \textbf{Multimodality of V2G.}  
Another advantage of GeoMan's image-to-video formulation is its ability to share model weights between the normal and depth prediction by simply switching the reference image. This enables training on multi-modal data, improving both generalization and training efficiency. Table~\ref{tab:ablation}(c) shows that the multi-modal model achieves superior results compared to task specific models, demonstrating the effectiveness of our formulation. 

\noindent \textbf{Source of the First Frame for V2G.}  
Tab.~\ref{tab:ablation}(d) shows that replacing I2G predictions with ground-truth normal or depth improves V2G's performance, highlighting the potential for further gains by refining the image-based I2G model.

\vspace{-1mm}
\section{Conclusion}
\vspace{-2mm}
We introduce GeoMan, a novel framework for accurate and temporally consistent depth and normal estimation from monocular videos. GeoMan addresses the challenge of 4D human data scarcity by framing the video geometry estimation task as image-to-video synthesis. To preserve human scale information, we proposed a root-relative depth representation tailored for human geometry. Extensive experiments on high-resolution benchmarks showed that GeoMan surpasses existing methods, even those trained on large-scale proprietary data, demonstrating effectiveness in real-world video geometry estimation and 3D reconstruction.

{
    \small
    \bibliographystyle{ieeenat_fullname}
    \bibliography{main}
}

\clearpage

\appendix

\onecolumn
\renewcommand{\theequation}{S\arabic{equation}}
\renewcommand{\thefigure}{S\arabic{figure}}
\renewcommand{\thetable}{S\arabic{table}}

\begin{center}
\bigskip 
\bigskip 
\Large{
\textbf{Supplementary Material of \\ GeoMan: Temporally Consistent Human Geometry Estimation \\ using Image-to-Video Diffusion}}
\bigskip 
\bigskip 
\end{center}%

In this documents, we provide additional results and details. We highly encourage readers to view our supplementary video introduced in \href{https://research.nvidia.com/labs/dair/geoman}{our project page}. We include more experimental and evaluation details in Sec.~\ref{sec:supp_imp_details}. Additional results, including geometry estimation on long videos and multi-person videos, quantitative evaluation on the Goliath dataset~\cite{martinez2024codec} and more ablation studies are present in Sec.~\ref{sec:supp_add_res}. Finally, we discuss the limitations of GeoMan and future works in Sec.~\ref{sec:supp_limitation}.

\section{Experimental Details}
\label{sec:supp_imp_details}
\subsection{Implementation Details}

We implemented the Image Geometry Diffusion (I2G) model based on the pretrained weights of Stable Diffusion 2 and the Video Geometry Diffusion (V2G) model using the pretrained weights of I2VGen-XL~\cite{zhang2023i2vgen}, initializing Video ControlNet with CtrlAdapter~\cite{lin2024ctrl}, and leveraging the \texttt{diffusers} library~\cite{von-platen-etal-2022-diffusers}.  

As shown in Fig. 2 of the main paper, the VAE encoder-decoder and Video ControlNet remain frozen, while only the diffusion denoiser, V2G, is fine-tuned.

Note that we learn separate diffusion models for image-based normal and depth map estimation. This is because predicting depth and normal from RGB images requires a diffusion model to map an RGB image to two significantly different outputs, thus separate models outperform a single model. 

Both models are trained at \(512 \times 512\) resolution using the Adam optimizer~\cite{kingma2014adam} with a learning rate of \(1e{-5}\). I2G is trained for 20K iterations with a batch size of 144, taking approximately two days on 4 NVIDIA A100 GPUs. V2G is trained on 12-frame sequences for 30K iterations with a batch size of 8, requiring approximately three days on 8 A100 GPUs. Evaluation is performed on 32-frame inputs for moving subject videos and 16-frame inputs for moving camera videos. For all experiments, we set the number of denoising steps to 100 and use an ensemble size of 8.

We preprocess input images by removing backgrounds, cropping the human region, and resizing images to a resolution of  $512\times512$. For inference on in-the-wild videos, we employ video matting method, BiRefNet~\cite{zheng2024birefnet}, to remove backgrounds. This preprocessing ensures that our model focuses solely on human-centric geometry while maintaining robustness across diverse scenarios.

\subsection{Evaluation Details}
\noindent \textbf{Evaluation Dataset}
The original ActorsHQ dataset includes eight actors performing 15 sequences, captured with 160 cameras at 25 fps.
For the evaluation, we generate 192 videos of moving subjects with static cameras (eight actors, 24 cameras, 32 frames per video) at various body scales. 
we also construct 256 videos with moving cameras and static subjects (32 cameras, 16 frames per video). 

\noindent \textbf{Depth Alignment}
For 'Optimizing shift' in the comparison of depth estimation methods,  we optimized the shift per frame with the ground truth depth maps . For 'Optimizing scale + shift,' we optimized the scale per sequence following ~\cite{hu2024-DepthCrafter} and optimized the shift per frame.

\noindent \textbf{Point Cloud Reconstruction}
Root depth is estimated using a 3D pose estimation method~\cite{wang2024tram} to convert predictions into metric depth. Point cloud reconstruction is performed using the dataset's camera intrinsic parameters. While we present unaligned reconstructions, we also align depth predictions to ground truth using an optimized shift to highlight scale differences, as shown in Figs.~\ref{fig:depth_supp1}–\ref{fig:depth_supp4}.

\noindent \textbf{Fine-tuning Sapiens on Our Training Dataset}
For the fine-tuning of Sapiens*, we use the pretrained Sapiens-2B model, which serves as the foundation model trained on a large-scale human dataset. We fine-tune this model specifically for the normal and root-relative depth  estimation using our training dataset, following their fine-tuning pipeline \footnote{\url{https://github.com/facebookresearch/sapiens}}.

\subsection{Metrics}

For evaluating the predicted depth, we follow \cite{eigen2014depth} and use the following metrics:

\begin{equation}\label{eqn:metrics}
\begin{array}{l}
\text{Abs Relative: } \frac{1}{\sum(K_t == 1)} \sum_{k_t \in K, d_t \in D_t} k_t \left\|\frac{d_t - d_t^{gt}}{d_t^{gt}}\right\| \\
\text{Squared Relative: } \frac{1}{\sum(K_t == 1)} \sum_{k_t \in K, d_t \in D_t} \frac{\left\|d_t - d_t^{gt}\right\|^2}{d_t^{gt}} \\
\operatorname{RMSE~(linear): } \sqrt{\frac{1}{\sum(K_t == 1)} \sum_{k_t \in K, d_t \in D_t} \left\|d_t - d_t^{gt}\right\|^2} \\
\operatorname{RMSE~(log): } \sqrt{\frac{1}{\sum(K_t == 1)} \sum_{k_t \in K, d_t \in D_t} \left\|\log d_t - \log d_t^{gt}\right\|^2} \\
\delta < \text{thr: } \frac{1}{\sum(K_t == 1)} \sum_{k_t \in K, d_t \in D_t} K_t \left[\text{Max}\left(\frac{D_t}{D_t^{gt}}, \frac{D_t^{gt}}{D_t}\right) < \text{thr}\right] \\
\delta 1 \text{: } \frac{1}{\sum(K_t == 1)} \sum_{k_t \in K, d_t \in D_t} K_t \left[\text{Max}\left(\frac{D_t}{D_t^{gt}}, \frac{D_t^{gt}}{D_t}\right) < \text{1.25}\right] \\
\text{SI-log10: } \sqrt{\frac{1}{\sum(K_t == 1)} \sum_{k_t \in K, d_t \in D_t} \left\|\log_{10}(d_t) - \log_{10}(d_t^{gt})\right\|^2}
\end{array}
\end{equation}
where $K_t$ is a depth validity mask, $D_t$ is the predicted depth for image $I_t$, and $D_t^{gt}$ is the ground-truth depth.

For evaluating the predicted normal maps, we use the metrics from Sapiens. The angular error is computed as the mean and median angular errors between predicted normal vectors and ground-truth normal vectors:

\begin{equation}
\text{Angular Error} = \frac{1}{n} \sum_{i=1}^{n} \cos^{-1} \left( \frac{{\mathbf{n}}_i \cdot \mathbf{n}_i^{gt}}{|{\mathbf{n}}_i| |\mathbf{n}_i^{gt}|} \right)
\end{equation}

Additionally, we compute the percentage of pixels where the angular error is below specific thresholds:
\begin{equation}
\text{Percentage of Pixels with Angular Error} < t^\circ = \frac{1}{n} \sum_{i=1}^{n} \mathbb{I}\left( \cos^{-1} \left( \frac{{\mathbf{n}}_i \cdot \mathbf{n}_i^{gt}}{|{\mathbf{n}}_i| |\mathbf{n}_i^{gt}|} \right) < t \right)
\end{equation}
where \( t \in \{11.5^\circ, 20^\circ\} \).

For evaluating temporal consistency, we introduce optical flow-based metrics. The optical flow-based warping metric (OPW) is defined by \cite{wang2022less} as:

\begin{equation}
OPW = \frac{1}{n} \sum_{i=1}^{n} \sum_{t=0}^{T-1} W_{t+1 \Rightarrow t}^{(i)} \left\| D_{t+1}^{(i)} - \hat{D}_t^{(i)} \right\|_1,
\end{equation}
where $W_{t+1 \Rightarrow t}^{(i)}$ is the optical flow-based visibility mask calculated from the warping discrepancy between subsequent frames, as explained in \cite{wang2022less}. We use the optical flow generated by the latest SOTA FlowFormer \cite{huang2022flowformer}. This metric is applied to both depth and normal frames, where the depth and normal values are warped between consecutive frames.

To further encourage comprehensive evaluation of temporal consistency, we introduce additional metrics. For normal frames, we use the following:

\begin{equation}
\text{TC-Mean (Optical Flow-based Angular Error)} = \frac{1}{n} \sum_{i=1}^{n} \frac{1}{T-1} \sum_{t=0}^{T-1} W_{t+1 \Rightarrow t}^{(i)} \cos^{-1} \left( \frac{\hat{\mathbf{n}}_t^{(i)} \cdot {\mathbf{n}}_{t+1}^{(i)}}{|\hat{\mathbf{n}}_t^{(i)}| |{\mathbf{n}}_{t+1}^{(i)}|} \right)
\end{equation}

\begin{equation}
\text{TC-}11.25^\circ \text{ (Optical Flow-based Angular Error)} = \frac{1}{n} \sum_{i=1}^{n} \frac{1}{T-1} \sum_{t=0}^{T-1} W_{t+1 \Rightarrow t}^{(i)} \mathbb{I} \left( \cos^{-1} \left( \frac{\hat{\mathbf{n}}_t^{(i)} \cdot {\mathbf{n}}_{t+1}^{(i)}}{|\hat{\mathbf{n}}_t^{(i)}| |{\mathbf{n}}_{t+1}^{(i)}|} \right) < 11.25^\circ \right)
\end{equation}
where \( \hat{\mathbf{n}}_t^{(i)} \) is the warped normal for the \( i \)-th sample at frame \( t \), and $ \mathbf{n}_{t+1}^{(i)} $ is the depth from the next frame.

For depth frames, we define the following temporal consistency metric:

\begin{equation}
\text{TC-RMSE (Optical Flow-based Temporal Consistency)} = \sqrt{\frac{1}{n} \sum_{i=1}^{n} \frac{1}{T-1} \sum_{t=0}^{T-1} W_{t+1 \Rightarrow t}^{(i)} \left\|\hat{D}_t^{(i)} - D_{t+1}^{(i)}\right\|^2}
\end{equation}

\begin{equation}
\text{TC-}\delta_1 \text{ (Optical Flow-based Depth Consistency)} = \frac{1}{n} \sum_{i=1}^{n} \frac{1}{T-1} \sum_{t=0}^{T-1} W_{t+1 \Rightarrow t}^{(i)} \mathbb{I} \left( \max \left( \frac{\hat{D}_t^{(i)}}{D_{t+1}^{(i)}}, \frac{D_{t+1}^{(i)}}{\hat{D}_t^{(i)}} \right) < 1.25 \right)
\end{equation}
where \( \hat{D}_t^{(i)} \) is the warped depth for the \( i \)-th sample at frame \( t \), and \( D_{t+1}^{(i)} \) is the depth from the next frame.

\section{Additional Qualitative and Quantitative Results}
\label{sec:supp_add_res}
\subsection{Geometry Estimation on Long Videos and Multi-Person Videos}

\begin{figure*}[!t]
    \centering
    \includegraphics[width=0.95\textwidth]{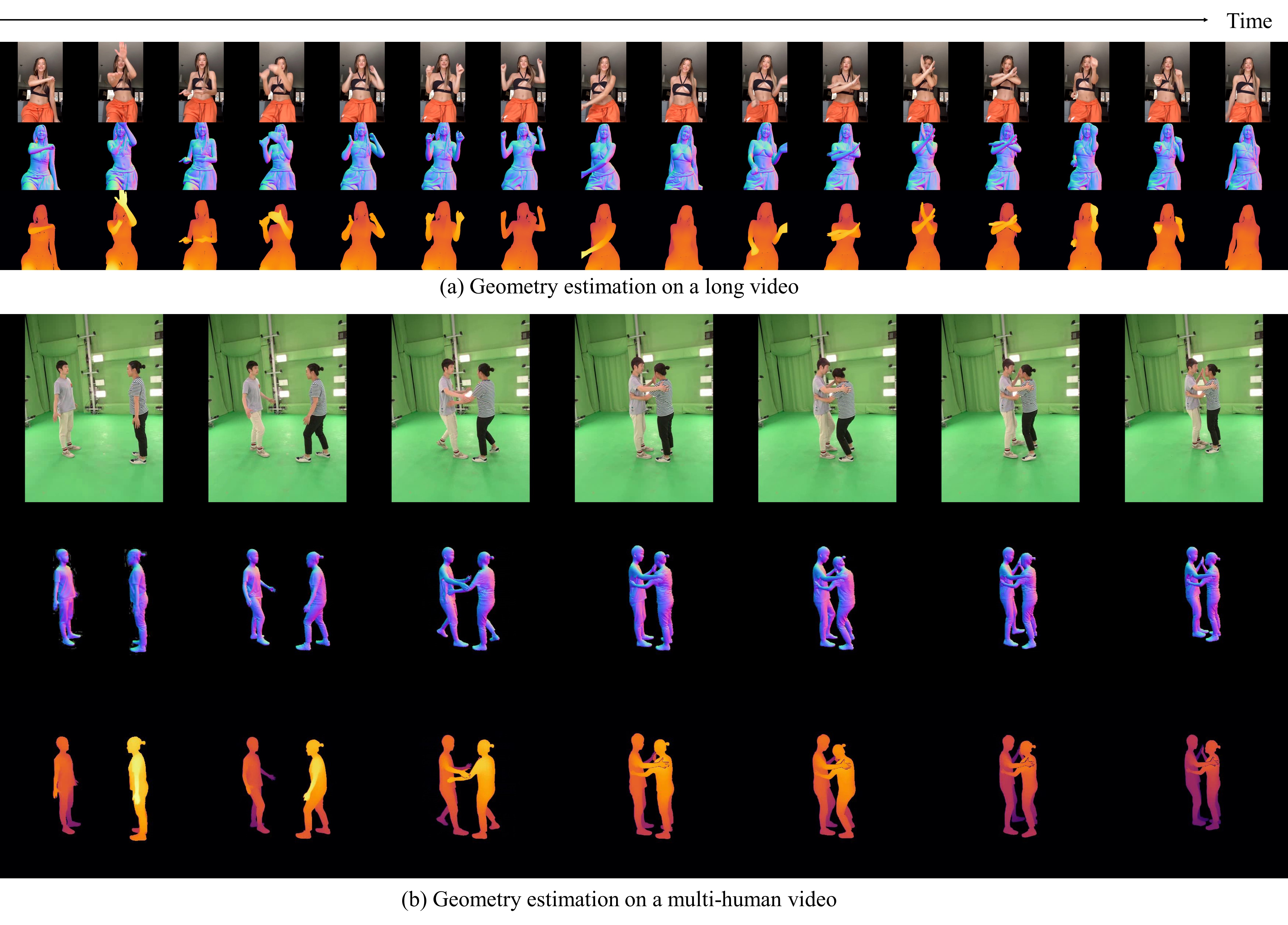}
    \vspace{-1.em}
    \caption{Results of geometry estimation on long and multi-person videos.}
    \label{fig_long_multi}
\end{figure*}

Our model effectively generalizes to long video sequences and multi-person scenarios (Fig.~\ref{fig_long_multi}). Despite being trained on fixed-length sequences, it supports inference on 64-frame sequences using an NVIDIA A100, leveraging robust temporal modeling and extending further via autoregressive inference.

To ensure smooth transitions across segments, we adopt the mortise-and-tenon-style latent interpolation from~\cite{hu2024-DepthCrafter}. Overlapping frames between segments are interpolated with linearly decreasing weights, and the final depth and normal sequences are reconstructed by decoding the stitched latent representations via the VAE decoder.

For multi-person scenarios, our approach remains effective despite training on single-person data. Normal estimation requires no modification, as pretrained priors enable natural adaptation. For depth estimation, we address root-relative ambiguity via per-subject masking and result aggregation. Human masks segment individuals, root-relative depths are estimated, and metric depths are reconstructed via 3D pose estimation, ensuring globally consistent depth while preserving per-subject geometric integrity.

\subsection{Qualitative Results on In-the-Wild Videos} 
\begin{figure*}[!h]
    \centering
    \includegraphics[width=1\textwidth]{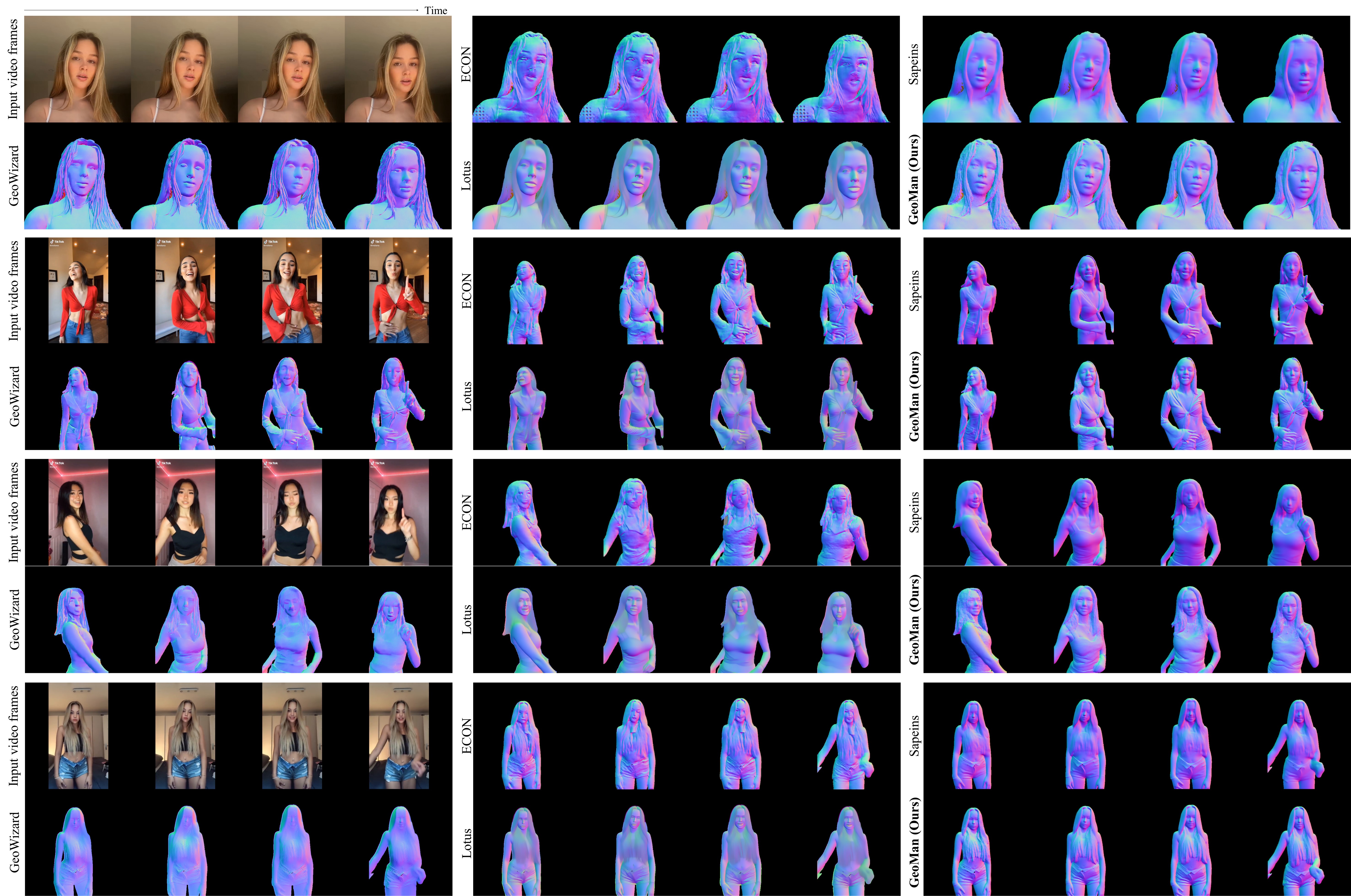}
    \caption{Comparison on zero-shot normal estimation results on in-the-wild videos.}
    \label{fig_supp_tiktok_normal}
\end{figure*}

\begin{figure*}[!h]
    \centering
    \includegraphics[width=1\textwidth]{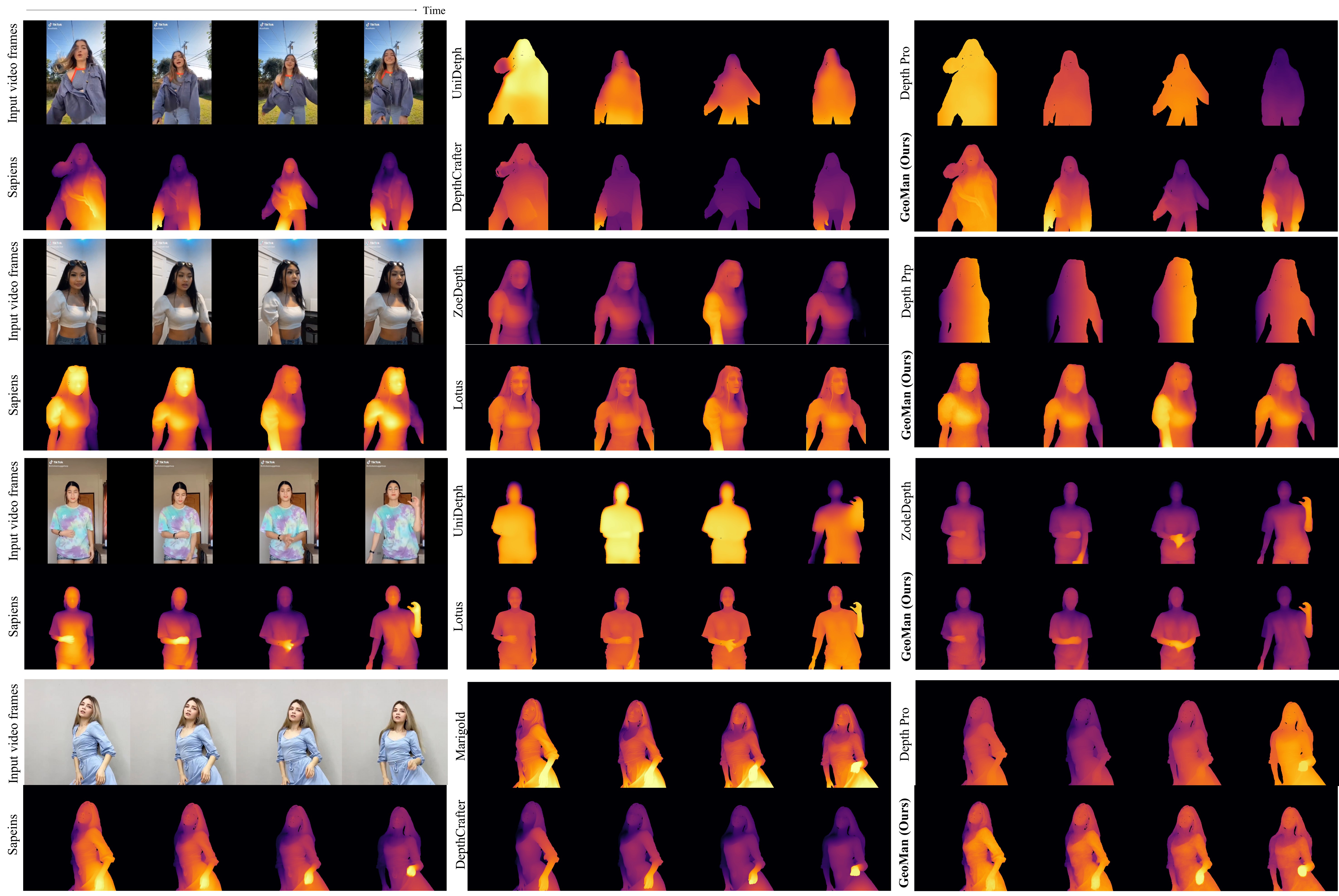}
    \caption{Comparison on zero-shot depth estimation results on in-the-wild videos.}
    \label{fig_supp_tiktok_depth}
\end{figure*}

To evaluate generalization to in-the-wild videos, we perform a qualitative comparison on short-form videos collected in Champ~\cite{zhu2024champ}, which compiles content from platforms such as Bilibili, Kuaishou, TikTok, YouTube, and Xiaohongshu. This dataset includes a diverse range of individuals in full-body, half-body, and close-up shots, with varying ages, ethnicities, genders, and backdrops.

As shown in Fig.~\ref{fig_supp_tiktok_normal} and~\ref{fig_supp_tiktok_depth}, our results demonstrate robust performance and high-quality geometry in these in-the-wild videos.

\subsection{Quantitative Results on Goliath Dataset}

\begin{table*}[!h]\centering
\scriptsize
\vspace{-1em}

\centering
\begin{adjustbox}{width=0.8\linewidth}
\begin{tabular}{lccccccccc}\toprule
&\multicolumn{6}{c}{\textbf{Performance on moving subject videos}} &\multicolumn{3}{c}{\multirow{2}{*}{\textbf{Temporal consistency}}} \\\cmidrule{2-7}
&\multicolumn{3}{c}{Optimizing shift} &\multicolumn{3}{c}{Optimizing scale + shift} & & & \\\cmidrule{2-10}
&AbsRel↓ &$\delta<$1.05↑ &$\delta$1↑ &AbsRel↓ &$\delta<$1.05↑ &$\delta$1↑ &OPW↓ &TC-RMSE↓ &TC-$\delta$1↑ \\\midrule
Sapiens &0.030 &0.807 &\textbf{1.000} &0.007 &0.994 &\textbf{1.000} &0.077 &0.024 &0.754 \\
Sapiens* &0.017 &0.954 &\textbf{1.000} &0.011 &0.986 &\textbf{1.000} &0.064 &0.022 &0.855 \\
Marigold &0.022 &0.900 &\textbf{1.000} &0.020 &0.904 &\textbf{1.000} &0.098 &0.031 &0.605 \\
GeoWizard &0.018 &0.933 &\textbf{1.000} &0.015 &0.955 &0.999 &0.061 &0.024 &0.709 \\
Lotus &0.023 &0.888 &\textbf{1.000} &0.022 &0.898 &\textbf{1.000} &0.097 &0.031 &0.680 \\
Depthcrafter &\ul{0.015} &0.957 &\textbf{1.000} &0.011 &0.985 &\textbf{1.000} &\ul{0.039} &\ul{0.019} &\textbf{0.919} \\
Depth any video &\ul{0.015} &\ul{0.960} &\textbf{1.000} &0.013 &0.979 &0.999 &0.044 &0.020 &0.866 \\
\textbf{GeoMan (Ours)} &\textbf{0.011} &\textbf{0.987} &\textbf{1.000} &\ul{0.010} &\ul{0.989} &\textbf{1.000} &\textbf{0.038} &\textbf{0.018} &\ul{0.909} \\
\bottomrule
\end{tabular}
\end{adjustbox}
\caption{Comparison with  affine-invariant depth estimation methods on zero-shot settings on Goliath. Sapiens*: Finetuned on our dataset.}
\label{tab_quanti_goliath2}
\end{table*}

We provide additional quantitative results on the Goliath~\cite{martinez2024codec} dataset, a multiview video dataset designed for studying complete avatars, including full-body geometry and underlying body shape. Each instance consists of eight captures of the same person, with four sequences featuring four subjects in full-body capture. For our evaluation, we used 10 views with 32 frames as the test set.

\begin{table*}[!h]\centering
\scriptsize
\vspace{-1em}

\centering
\begin{adjustbox}{width=0.7\linewidth}
\begin{tabular}{lcccccccc}\toprule
&\multicolumn{4}{c}{\textbf{Performance on moving subject videos}} &\multicolumn{3}{c}{\textbf{Temporal consistency}} \\\cmidrule{2-8}
&Mean↓ &Median↓ &11.25°↑ &30°↑ &OPW↓ &TC-Mean↓ &TC-11.25°↑ \\\cmidrule{2-8}
ECON &25.498 &20.129 &21.800 &72.345 &0.196 &16.280 &53.385 \\\midrule
Sapiens &\textbf{11.205} &\textbf{8.856} &\textbf{64.284} &\textbf{96.091} &\ul{0.104} &\ul{9.653} &\ul{77.615} \\
Sapiens* &13.531 &10.605 &53.533 &93.712 &0.113 &10.812 &74.594 \\
GeoWizard &19.622 &16.167 &30.789 &82.829 &0.190 &18.037 &41.474 \\
Lotus &17.999 &14.766 &35.198 &86.666 &0.104 &10.155 &75.924 \\
ICLight &27.664 &24.571 &14.084 &63.199 &0.169 &14.513 &56.882 \\
GeoMan (Ours) &\ul{12.831} &\ul{10.034} &\ul{56.665} &\ul{94.308} &\textbf{0.084} &\textbf{8.128} &\textbf{82.811} \\
\bottomrule
\end{tabular}
\end{adjustbox}
\caption{Comparison on zero-shot normal estimation on Goliath. Sapiens*: Fine-tuned on our dataset.}
\label{tab:tab_quanti_goliath1}
\end{table*}

As shown in Table~\ref{tab:tab_quanti_goliath1} and \ref{tab_quanti_goliath2}, while GeoMan significantly outperforms other methods, the pre-trained version of Sapiens~\cite{khirodkar2024sapiens} remains competitive. However, it is important to note that Sapiens was trained on a proprietary dataset. To ensure a fair comparison, we fine-tuned Sapiens's pre-trained MAE model on our dataset. Under this fair setting, GeoMan outperforms Sapiens by a notable margin.

\subsection{Additional Results on ActorsHQ} 
\begin{figure*}[!h]
    \centering
    \includegraphics[width=1\textwidth]{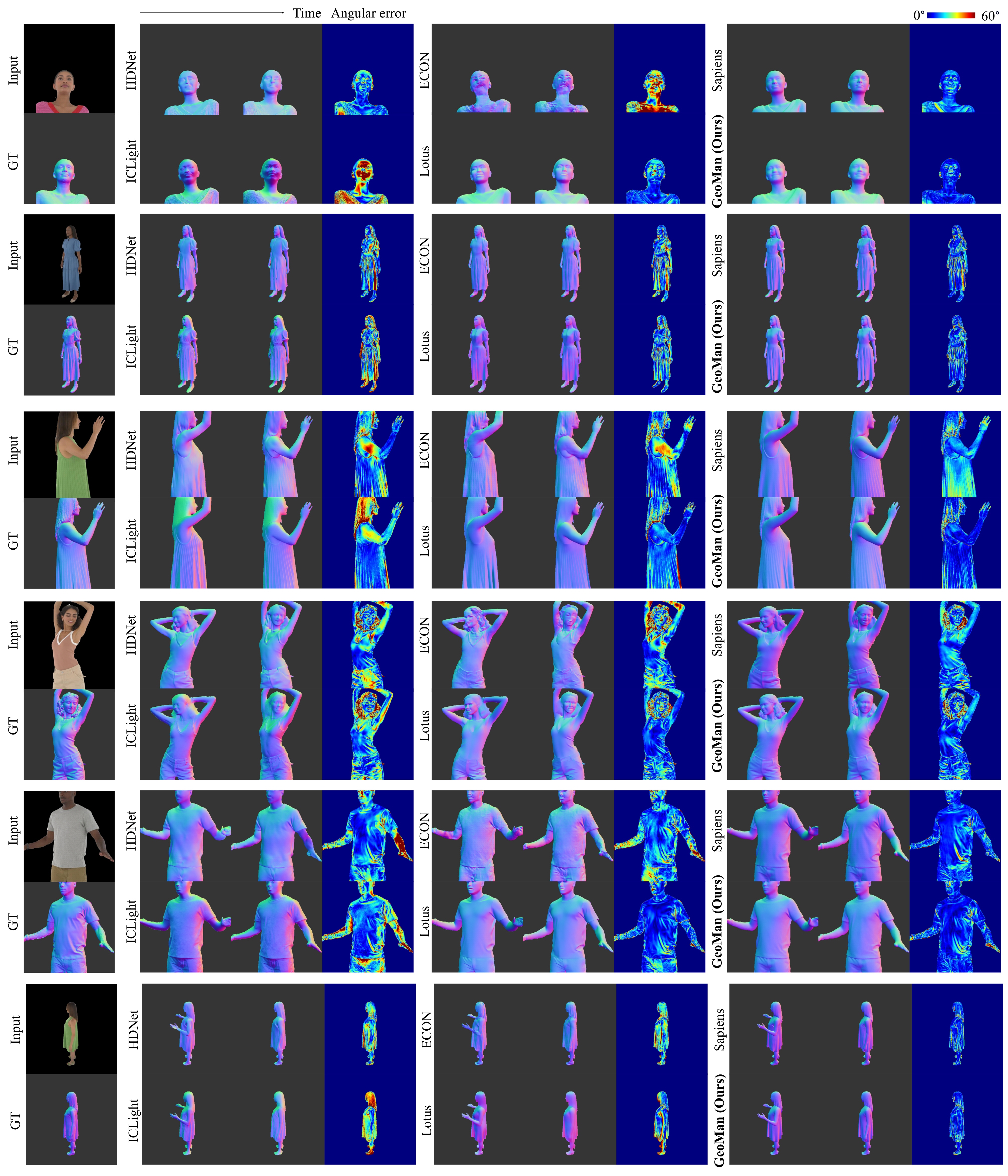}
    \caption{Comparison on zero-shot normal estimation on ActorsHQ. \textbf{Left}: Predicted normal. \textbf{Right}: Visualization of angular error.}
    \label{fig:normal_supp1}
\end{figure*}

Fig.~\ref{fig:normal_supp1} presents additional qualitative comparisons in surface normal estimation, demonstrating that our method not only achieves superior accuracy over baselines but also ensures greater temporal consistency and higher fidelity.

\begin{figure*}[!h]
    \centering
    \includegraphics[width=0.9\textwidth]{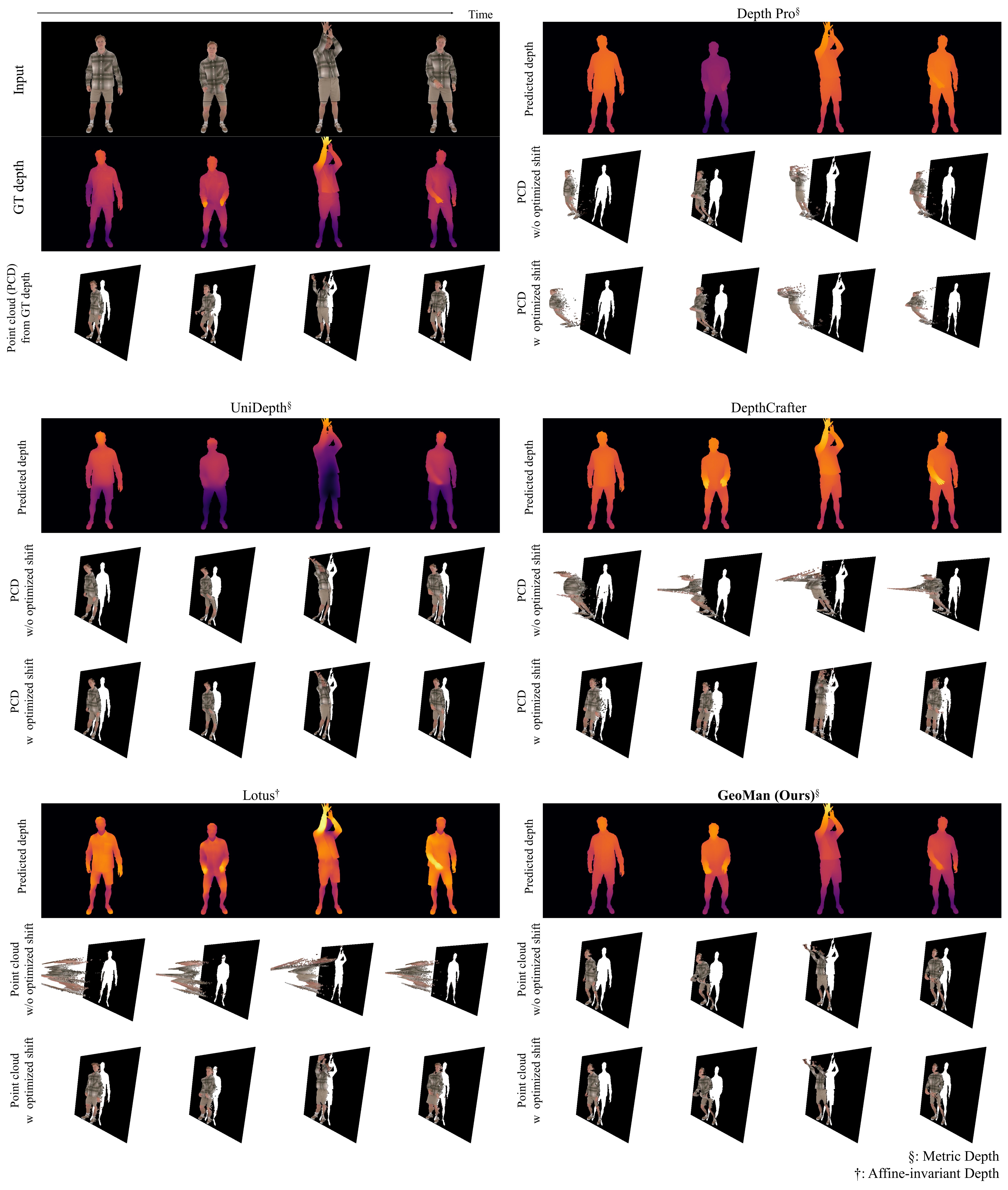}
     \caption{
    Comparison with the existing depth estimation models.
    GeoMan produces the state-of-the-art results in both temporal consistency and fidelity.
    }
    \label{fig:depth_supp1}
\end{figure*}

\begin{figure*}[!h]
    \centering
    \includegraphics[width=0.9\textwidth]{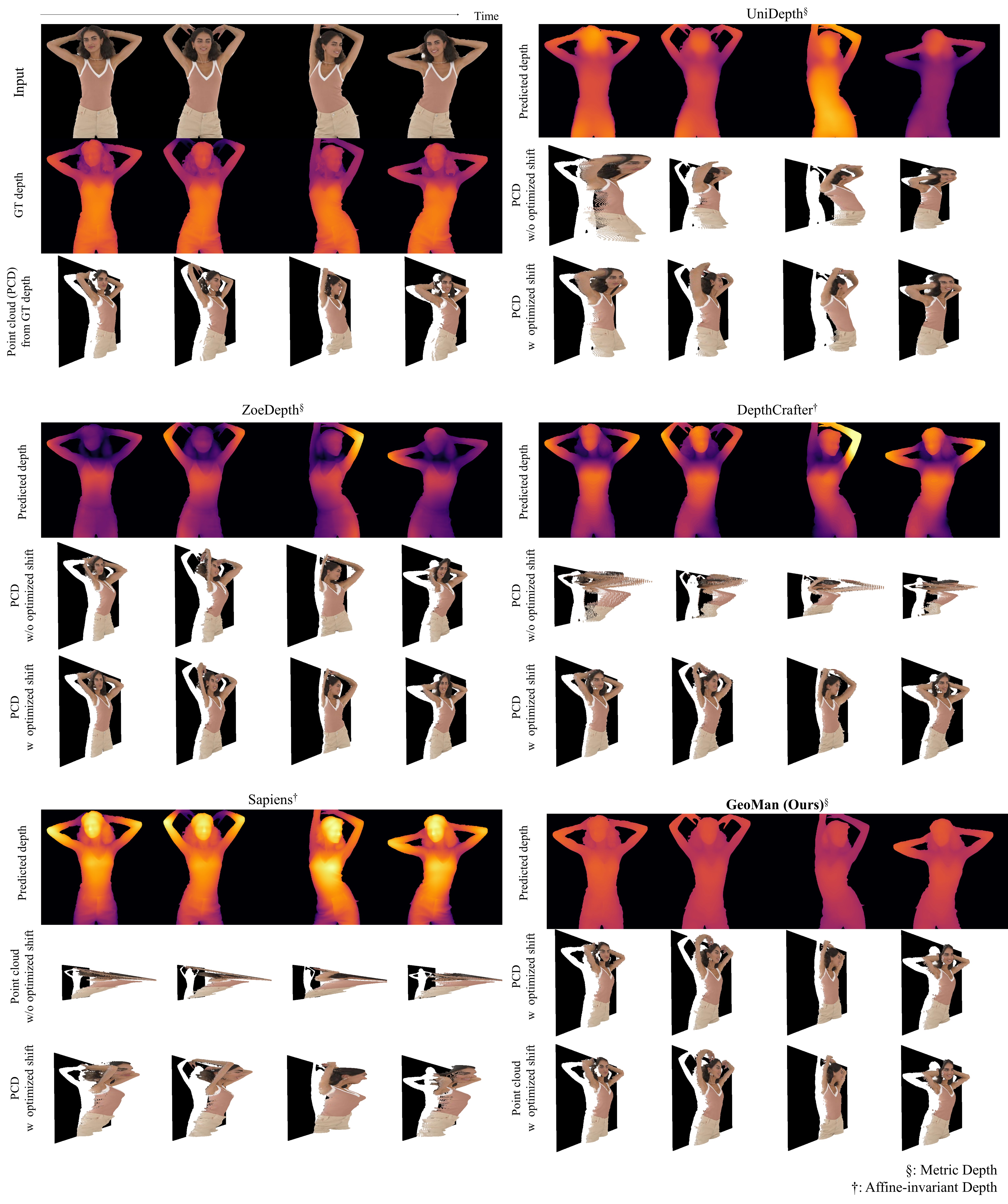}
     \caption{
    Comparison with the existing depth estimation models.
    GeoMan produces the state-of-the-art results in both temporal consistency and fidelity.
    }
    \label{fig:depth_supp2}
\end{figure*}

\begin{figure*}[!h]
    \centering
    \includegraphics[width=0.9\textwidth]{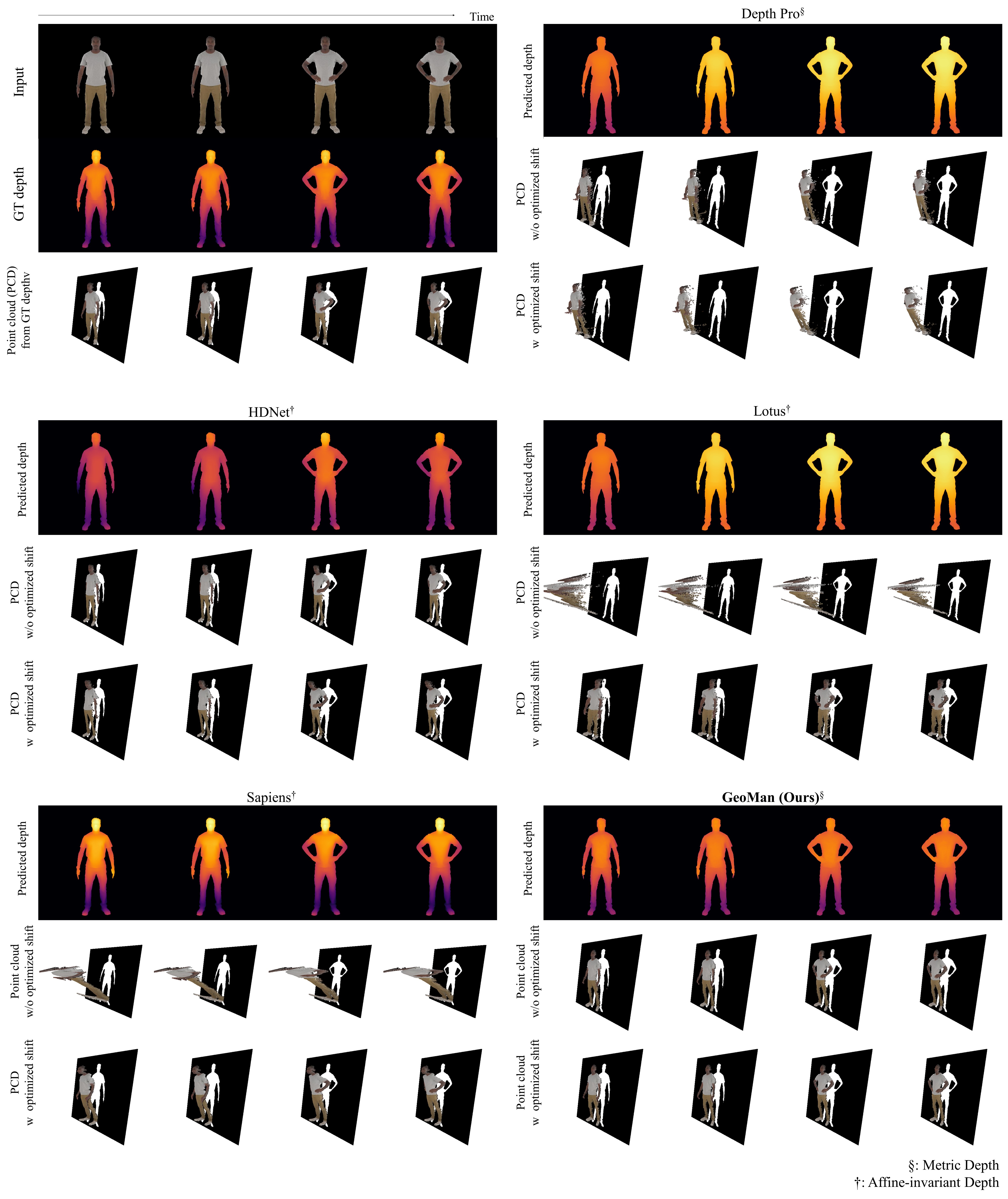}
     \caption{
    Comparison with the existing depth estimation models.
    GeoMan produces the state-of-the-art results in both temporal consistency and fidelity.
    }
    \label{fig:depth_supp3}
\end{figure*}

\begin{figure*}[!h]
    \centering
    \includegraphics[width=0.9\textwidth]{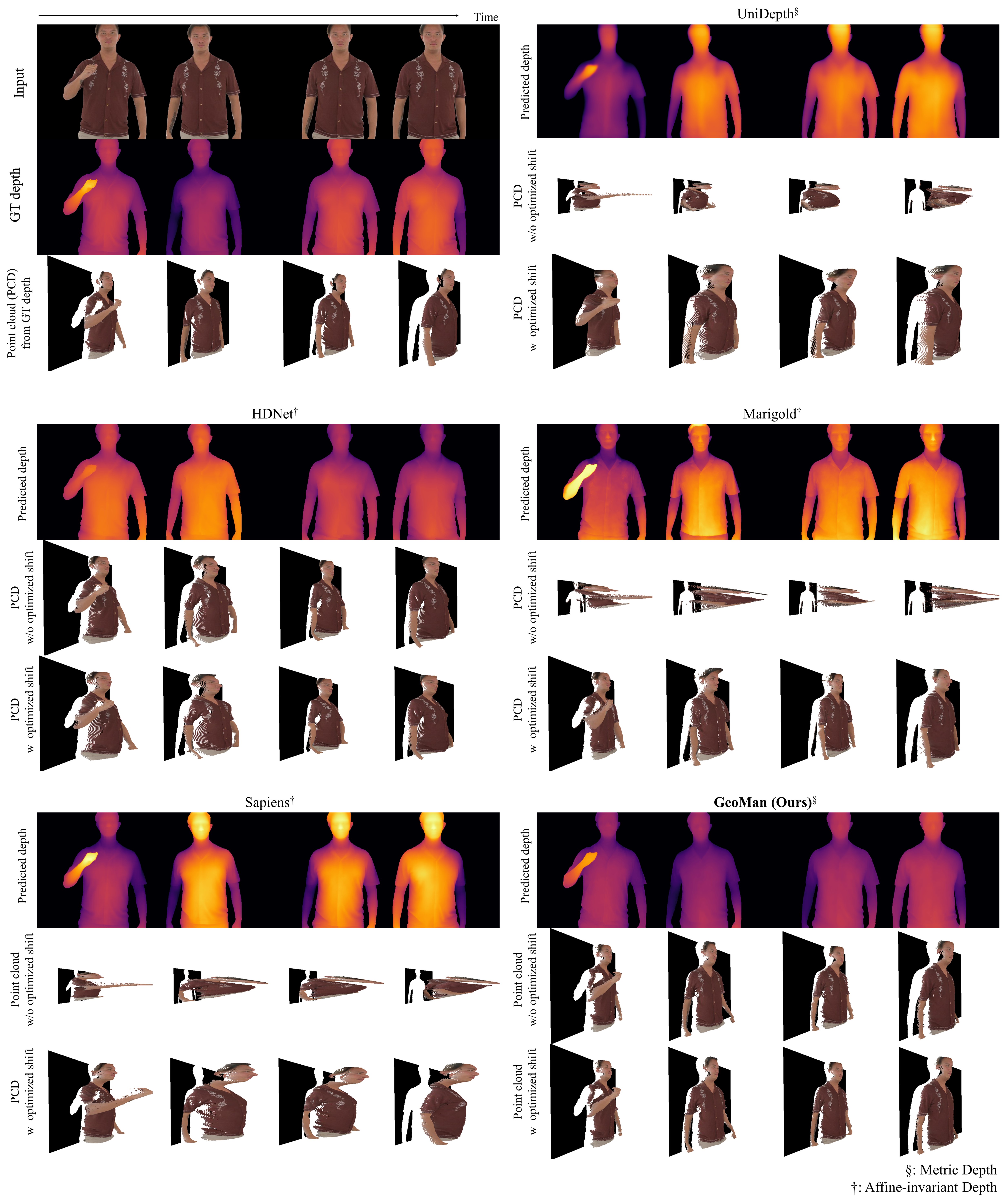}
     \caption{
    Comparison with the existing depth estimation models.
    GeoMan produces the state-of-the-art results in both temporal consistency and fidelity.
    }
    \label{fig:depth_supp4}
\end{figure*}

\begin{figure*}[!h]
    \centering
    \includegraphics[width=0.9\textwidth]{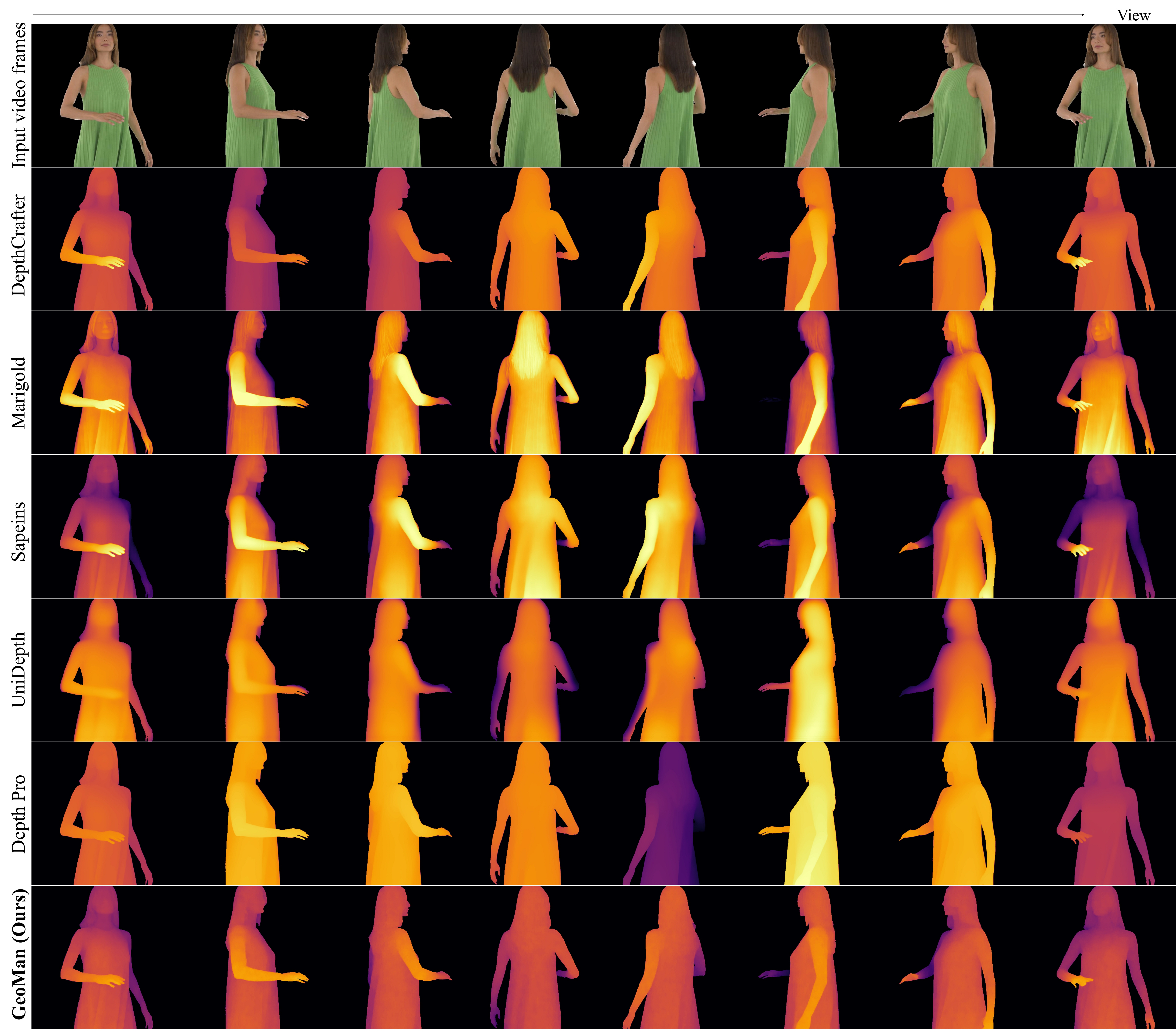}
     \caption{
    Comparison with the existing depth estimation models.
    GeoMan produces the state-of-the-art results in both temporal consistency and fidelity.
    }
    \label{fig:depth_supp5}
\end{figure*}

\begin{figure*}[!h]
    \centering
    \includegraphics[width=0.9\textwidth]{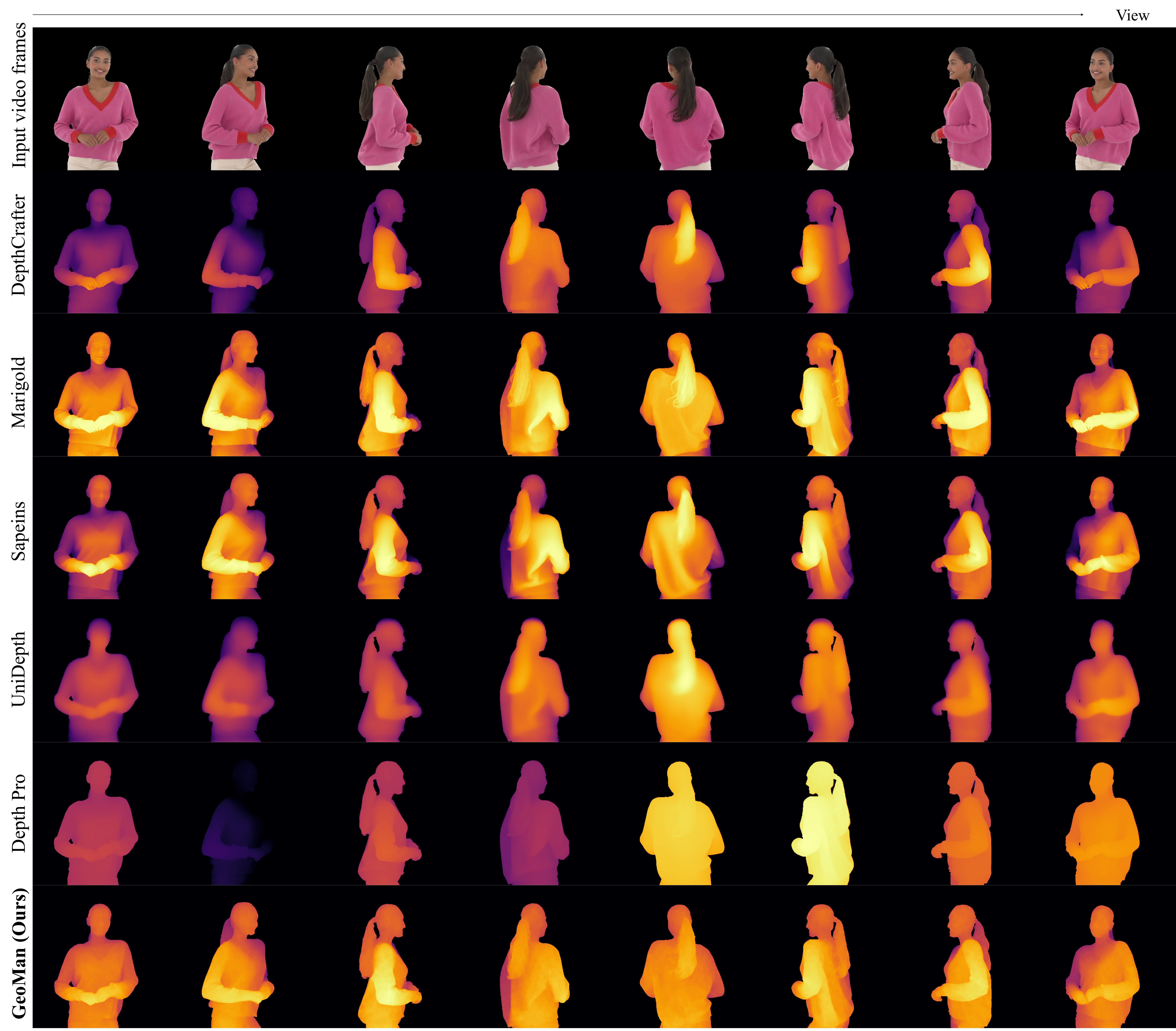}
     \caption{
    Comparison with the existing depth estimation models.
    GeoMan produces the state-of-the-art results in both temporal consistency and fidelity.
    }
    \label{fig:depth_supp6}
\end{figure*}

Figs.~\ref{fig:depth_supp1}–\ref{fig:depth_supp6} provide further qualitative comparisons with baseline methods forboth moving camera and moving subject videos. To ensure consistency, all predicted depth maps are renormalized using sequence-wise min-max scaling within the human mask. Compared to per-frame estimation models, our approach produces more temporally stable results.

For point cloud reconstruction, we present both unaligned reconstructions and those with depth aligned to ground truth using an optimized shift to emphasize scale differences. Our method effectively preserves human scale, demonstrating strong geometric consistency.

\subsection{Additional Ablation Studies and Analysis}

\begin{table*}[!t]\centering

\scriptsize
\label{tab:quanti}
\centering
\begin{adjustbox}{width=1\linewidth}
\begin{tabular}{lcccccccccccc}\toprule
&\multicolumn{5}{c}{\textbf{Normal}} &\multicolumn{6}{c}{\textbf{Depth}} \\\cmidrule{2-12}
&\multicolumn{2}{c}{Angular difference} &\%within t° &\multicolumn{2}{c}{Temporal consistency} &\multicolumn{2}{c}{Optimizing shift} &\multicolumn{2}{c}{Optimizing scale + shift} &\multicolumn{2}{c}{Temporal consistency} \\\cmidrule{2-12}
&Mean↓ &Median↓ &11.25°↑ &OPW↓ &TC-Mean↓ &AbsRel↓ &$\delta<$1.05↑ &AbsRel↓ &$\delta<$1.05↑ &OPW↓ &TC-RMSE↓ \\\midrule
&\multicolumn{11}{c}{\textbf{(a) Naïve extension vs GeoMan fine-tuning}} \\
Naïve-5K-steps &25.305 &22.732 &15.602 &0.076 &7.458 &0.038 &0.720 &0.028 &0.856 &0.049 &0.019 \\ 
Naïve-10K steps &23.550 &20.839 &18.601 &0.071 &6.938 &0.026 &0.866 &0.022 &0.907 &0.046 &0.017 \\
Naïve-20K steps &21.137 &18.435 &22.488 &0.070 &6.855 &0.021 &0.915 &0.018 &0.939 &0.040 &0.016 \\
Naïve-30K steps &19.669 &16.628 &28.405 &0.072 &7.169 &0.020 &0.926 &0.017 &0.943 &0.040 &0.016 \\
Naïve-50K steps &20.307 &17.131 &27.754 &0.075 &7.274 &0.017 &0.945 &0.016 &0.958 &0.041 &0.535 \\\arrayrulecolor{black!30}\midrule
\textbf{GeoMan-5K steps} &17.499 &13.541 &40.340 &0.069 &6.781 &0.014 &0.964 &0.013 &0.970 &0.036 &\textbf{0.014} \\
\textbf{GeoMan-10K steps} &17.255 &13.500 &40.094 &0.071 &6.881 &0.012 &0.975 &0.012 &0.981 &0.033 &\textbf{0.014} \\
\textbf{GeoMan-20K steps} &16.548 &12.719 &43.483 &\textbf{0.066} &\textbf{6.502} &\textbf{0.012} &0.977 &\textbf{0.011} &\textbf{0.984} &\textbf{0.032} &\textbf{0.014} \\
\textbf{GeoMan-30K steps} &16.185 &12.334 &45.217 &0.070 &6.876 &\textbf{0.012} &\textbf{0.978} &\textbf{0.011} &\textbf{0.984} &\textbf{0.032} &\textbf{0.014} \\
\textbf{GeoMan-50K steps} &\textbf{16.105} &\textbf{12.144} &\textbf{45.995} &0.067 &6.599 &\textbf{0.012} &0.975 &\textbf{0.011} &0.983 &0.033 &0.015 \\\arrayrulecolor{black}\midrule
& & & & & &\textbf{(b) I2G vs V2G} & & & & & \\
I2G &\textbf{16.033} &\textbf{12.068} &\textbf{46.222} &0.118 &11.261 &0.013 &0.970 &0.013 &0.974 &0.040 &0.016 \\
\textbf{I2G+V2G} &16.185 &12.334 &45.217 &\textbf{0.070} &\textbf{6.876} &\textbf{0.012} &\textbf{0.978} &\textbf{0.011} &\textbf{0.984} &\textbf{0.032} &\textbf{0.014} \\
\midrule
&\multicolumn{11}{c}{\textbf{ (c) Multimodality of I2G}} \\
Unimodal &\textbf{16.033} &\textbf{12.068} &\textbf{46.222} &0.118 &11.261 &0.013 &0.970 &0.013 &0.974 &0.040 &0.016 \\
\textbf{Multimodal} &18.442 &14.452 &36.985 &0.086 &9.930 &0.017 &0.946 &0.016 &0.956 &0.040 &0.017 \\\midrule
&\multicolumn{11}{c}{\textbf{(d) Training data source of V2G}} \\
Only 3D data &24.068 &20.184 &21.938 &0.107 &[9.966 &0.023 &0.883 &0.024 &0.883 &0.048 &0.017 \\
Only 4D data &16.387 &12.441 &44.787 &\textbf{0.062} &\textbf{6.198} &0.013 &0.967 &0.012 &0.980 &{0.035} &{0.015} \\
\textbf{Combined } &\textbf{16.185} &\textbf{12.334} &\textbf{45.217} &0.070 &6.876 &\textbf{0.012} &\textbf{0.978} &\textbf{0.011} &\textbf{0.984} &\textbf{0.032} &\textbf{0.014} \\
\midrule
&\multicolumn{11}{c}{\textbf{(e) Effectiveness of human area crop}} \\
w/o Human area crop &16.821 &12.891 &42.801 &\textbf{0.070} &6.974 &\textbf{0.012} &0.977 &0.012 &0.982 &0.037 &0.015 \\
\textbf{w Human area crop} &\textbf{16.185} &\textbf{12.334} &\textbf{45.217} &\textbf{0.070} &\textbf{6.876} &\textbf{0.012} &\textbf{0.978} &\textbf{0.011} &\textbf{0.984} &\textbf{0.032} &\textbf{0.014} \\
\midrule
&\multicolumn{11}{c}{\textbf{(f) Source of the First Frame for V2G}} \\
 1/4 Downscaled I2G &17.843 &13.510 &40.593 &0.072 &8.118 &0.016 &0.964 &0.015 &0.968 &0.040 &0.016 \\
\textbf{I2G} &16.185 &12.334 &45.217 &\textbf{0.070} &\textbf{6.876} &0.012 &0.978 &0.011 &0.984 &\textbf{0.032} &\textbf{0.014} \\
\bottomrule
\end{tabular}
\end{adjustbox}
\vspace{-0.5em}
\caption{Additional ablation studies. We validate the impact of various design choices.}
\label{tab:supp_ablation}
\end{table*}

\noindent \textbf{Naïve extension vs. GeoMan Fine-Tuning.}
In our experiments, we rigorously compare the performance of our proposed image-to-video formulation of GeoMan against a naïve extension of~\cite{marigold, hu2024-DepthCrafter}. Despite training for further 50K steps, the naïve extension strategy consistently yields suboptimal results across multiple metrics. As highlighted in Tab.~\ref{tab:supp_ablation}(a), the performance gap remains significant, reinforcing the superior efficiency and effectiveness of our GeoMan approach in handling complex video generation tasks. This stark contrast in performance emphasizes the importance of a more structured, model-specific fine-tuning strategy, which our GeoMan method embodies, leading to more robust and reliable results.

\begin{figure*}[!h]
    \centering
    \includegraphics[width=0.8\textwidth]{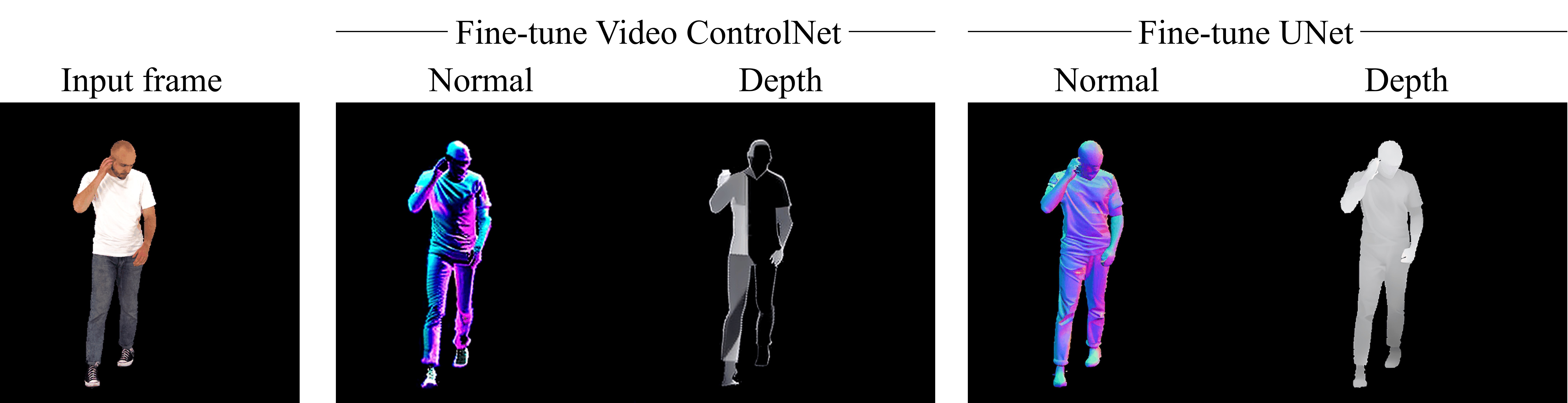}
    \vspace{-.5em}
    \caption{Fine-tuning the only Video ControlNet part is insufficient for reframing image-to-video diffusion models, whereas fine-tuning the UNet successfully transforms image-to-video generation into video geometry estimation.}
    \label{fig:ft_unet_controlnet}
\end{figure*}

\noindent \textbf{Fine-Tuning Video ControlNet vs. Fine-Tuning UNet.}
Further investigation into the finer details of our approach reveals a compelling insight into the critical components involved in fine-tuning. As illustrated in Fig.~\ref{fig:ft_unet_controlnet}, we demonstrate that fine-tuning solely the Video ControlNet component is not sufficient to effectively adapt image-to-video diffusion models to the geometry estimation method. In contrast, our approach of fine-tuning the UNet component proves far more effective, facilitating a seamless transition from image-based generation to accurate video geometry estimation. 

\noindent \textbf{I2G vs. V2G.}
We present a detailed comparison between the per-frame estimations of the image-to-geometry (I2G) and the full GeoMan pipeline in Tab.~\ref{tab:supp_ablation}(b). While I2G and GeoMan (I2G$+$V2G) exhibits similar accuracy, GeoMan outperforms I2G in temporal consistency. GeoMan pipeline excels in maintaining stability and coherence across frames, thereby ensuring smoother, more realistic video sequences. This superior temporal consistency is essential for high-quality video generation and demonstrates the power of GeoMan in addressing the challenges posed by dynamic visual content.

\noindent \textbf{Multimodality of I2G.}
I2G faces challenges in multimodal modeling, as shown in Tab.\ref{tab:supp_ablation}(c), requiring modifications to input and output layers, which limits performance. While the V2G framework, involving similar tasks in image-to-video translation, benefits from fine-tuning a single model, enhancing efficiency and performance, I2G's task divergence demands separate training for each modality.

\noindent \textbf{Training Data Sources for V2G.}  
In Tab.~\ref{tab:supp_ablation}(d), we evaluate the impact of different training datasets on V2G. Training only on rotating videos generated from static 3D scans introduces artifacts caused by motion bias. On the other hand, using only low-resolution 4D data results in a lack of fine details, limiting the quality of predictions. By combining both datasets, we achieve the best results, with improved visual quality and enhanced temporal consistency.

\noindent \textbf{Effectiveness of Human Area Crop.}
 As demonstrated in Tab.~\ref{tab:supp_ablation}(e), the use of human area cropping ensures that the model effectively targets relevant regions, thereby optimizing computational efficiency and accuracy. The positive impact of human area cropping confirms that focusing on key areas within the video frames is important, ensuring that unnecessary background noise does not interfere with the core task of human geometry estimation.

\noindent \textbf{First-Frame Dependency.}  
Replacing I2G predictions with ground truth improves performance as shown in Tab. 4 in the paper. Degrading the first frame by reducing its resolution by 4 times resulted in only a slight increase in error (Tab.~\ref{tab:supp_ablation}(f), indicating the robustness of our model. The degraded version still outperforms other baselines, including Sapiens.

\begin{figure}[h!]
\vspace{-5mm}
    \centering
    \includegraphics[width=0.9\textwidth]{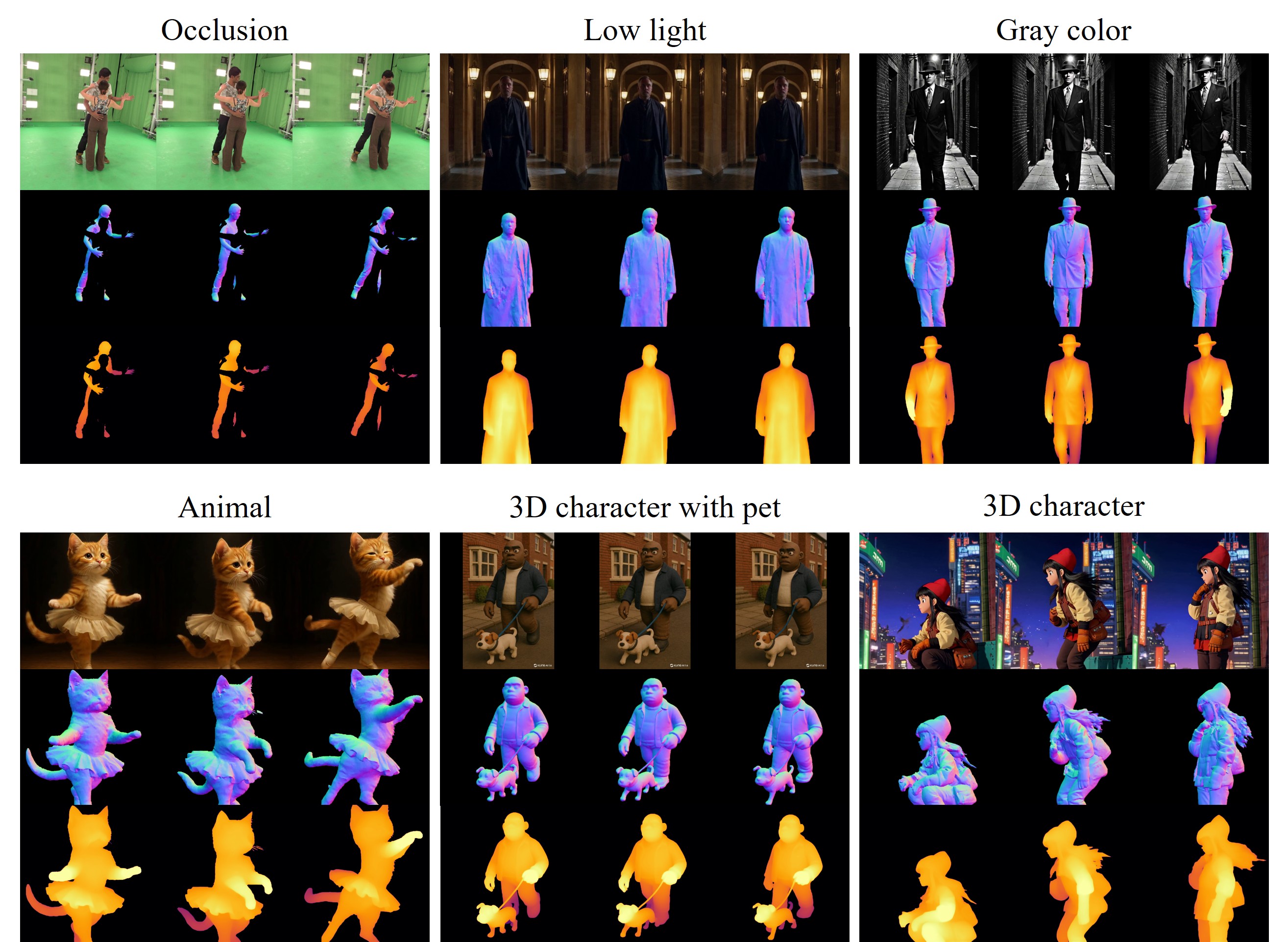}
     \caption{GeoMan demonstrates generalizablity across diverse scenarios, including occlusions, low lighting, and atypical human types.}
    \label{fig:generalization_rebuttal}
\end{figure}

\noindent \textbf{Robustness \& Generalizability.}  
We've shown generalization to long videos, multiple persons, and drastic poses in the supp. video. Fig.~\ref{fig:generalization_rebuttal} additionally highlights GeoMan’s robustness to diverse conditions, including occlusions, poor lighting, and atypical human types.

 \begin{table}[h!]
\centering
\begin{adjustbox}{width=0.5\linewidth}
\begin{tabular}{cccccc}
\hline
\textbf{Method} & \textcolor{gentlepink}{\textbf{GeoMan}} & \textcolor{gentlepink}{\textbf{Marigold}} & \textcolor{gentlepink}{\textbf{GeoWizard}} & \textcolor{gentlepink}{\textbf{IC-Light}} & \textcolor{gentleviolet}{\textbf{ECON}}\\
\hline
\textbf{Time (s)}  &10.57  & 8.44 & 23.34 & 7.854 & 134 \\
\hline
\textbf{Method} & \textcolor{gentleblue}{\textbf{Depthcrafter}} & \textcolor{gentleblue}{\textbf{Sapiens}}& \textcolor{gentleblue}{\textbf{Metric3Dv2}} & \textcolor{gentleblue}{\textbf{UniDepth}} & \textcolor{gentleblue}{\textbf{DepthPro}} \\
\hline
\textbf{Time (s)} & 3.59 & 1.92 & 0.31 & 0.56 & 0.90\\
\hline
\end{tabular}
\end{adjustbox}
\vspace{-3mm}
\caption{{Methods colored in \textcolor{gentlepink}{\textbf{red}} are diffusion-based, in \textcolor{gentleviolet}{\textbf{violet}} is optimization-based, in \textcolor{gentleblue}{\textbf{blue}} are feed-forward.}}
\label{tab:efficiency}
\end{table}

\noindent  \textbf{Efficiency Analysis.} As shown in Tab.~\ref{tab:efficiency}, GeoMan achieves competitive inference times compared to existing \textit{diffusion-based} methods, though it is slower than some other baselines. However, Geoman has clear benefits: significantly improved temporal consistency, richer 3D detail, and enhanced human-specific understanding.  
While inference time increases with ensemble size or the number of diffusion steps, this trade-off is adjustable and can be tuned per application. 
We envision GeoMan as a practical tool for acquiring large-scale, real-world supervision, enabling its distillation into the next generation of faster, lightweight human geometric estimation models.

\section{Limitations and Future Works}
\label{sec:supp_limitation}
For in-the-wild applications, our method relies on matting techniques to separate the subject from the background, making its performance inherently dependent on matting accuracy. Additionally, metric depth reconstruction is constrained by the precision of 3D human pose estimation, particularly the accuracy of pelvis root depth estimation.
Due to VRAM limitations, we trained our model at a maximum resolution of 512, fitting within 80GB of NVIDIA A100 GPUs with a batch size of 1. While our approach produces detailed and high-quality geometry, we plan to explore lightweight models or alternative training strategies to enable higher-resolution generation.

\end{document}